\pdfoutput=1

\documentclass[11pt]{article}

\usepackage{acl}
\usepackage{times}
\usepackage{latexsym}
\usepackage{graphicx}
\usepackage{tabularx}
\usepackage{lipsum}
\usepackage{multirow}
\usepackage{amssymb}
\usepackage{amsmath}
\usepackage{bm}
\usepackage{bbm}
\usepackage{makecell}
\usepackage{xcolor,colortbl}
\usepackage{subcaption}
\usepackage{hhline}
\usepackage[inline]{enumitem}
\usepackage[normalem]{ulem}
\usepackage[lined,ruled,algonl]{algorithm2e}
\usepackage{standalone}
\usepackage{todonotes}
\usepackage{tikz}
\usepackage{pgfplots}
\usepackage{xifthen}
\pgfplotsset{compat=1.16}
\usepgfplotslibrary{statistics}
\usepgflibrary{plotmarks}
\usetikzlibrary{calc, patterns}

\pgfplotsset{every axis/.append style={
    label style={font=\footnotesize},
    tick label style={font=\footnotesize},
    yticklabel style = {
        /pgf/number format/fixed,
        /pgf/number format/set thousands separator=,
        /pgf/number format/precision=5
    },
    scaled y ticks=false,
    xticklabel style = {
        /pgf/number format/fixed,
        /pgf/number format/set thousands separator=,
        /pgf/number format/precision=5
    },
    scaled x ticks=false,
}}

\newcommand{\tfigwidth}{9cm}
\newcommand{\tfigheight}{5cm}
\newcommand{\customlinewidth}{1.5pt}
\newcommand{\customlinewidthdotted}{2pt}
\newcommand{\customlegendxoffset}{12.6cm}
\newcommand{\customlegendyoffset}{4.6cm}
\newcommand{\lineopacity}{0.8}
\newcommand{\custompassivelinewidth}{0.8pt}

\newcommand{\cameraready}[1]{\textcolor{black}{#1}}

\definecolor{paultol1}{HTML}{4477AA}
\definecolor{paultol2}{HTML}{EE6677}
\definecolor{paultol3}{HTML}{228833}
\definecolor{paultol4}{HTML}{CCBB44}
\definecolor{paultol5}{HTML}{66CCEE}
\definecolor{paultol6}{HTML}{AA3377}
\definecolor{paultol7}{HTML}{BBBBBB}

\newcommand{\drawTrainingPlotSingleMetric}[9]{
    \begin{axis}[
        name={#8},
        at={#7},
        anchor={#9},
        title = {\textbf{\large #2}},
        xmin=0,
        max space between ticks=5000,
        try min ticks=5,
        label style={font=\large},
        tick label style = {font=\large},
        xlabel = {\# unique annotations},
        ylabel = {#3},
        height=\tfigheight,
        width=\tfigwidth,
        xmax={#4},
        legend style={
            at={(\customlegendxoffset,\customlegendyoffset)},
            anchor=north,
            /tikz/every even column/.append style={column sep=0.5cm},
            draw=none,
        },
        legend columns=11, 
        legend cell align=left,
    ]
    \addplot[
        paultol1,
        opacity=\lineopacity,
        line width=\customlinewidth,
    ] table [col sep=tab, y=random_active_y, x=random_active_x] {#1};

    \addplot[
        paultol1,
        dashed,
        opacity=\lineopacity,
        line width=\customlinewidthdotted,
    ] table [col sep=tab, y=uncertainty_active_y, x=uncertainty_active_x] {#1};

    \addplot[
        paultol2,
        opacity=\lineopacity,
        line width=\customlinewidth,
    ] table [col sep=tab, y=random_lmin_y, x=random_lmin_x] {#1};

    \addplot[
        paultol2,
        opacity=\lineopacity,
        dashed,
        line width=\customlinewidthdotted,
    ] table [col sep=tab, y=uncertainty_lmin_y, x=uncertainty_lmin_x] {#1};

    \addplot[
        paultol3,
        opacity=\lineopacity,
        line width=\customlinewidth,
    ] table [col sep=tab, y=random_semdiv_y, x=random_semdiv_x] {#1};

    \addplot[
        paultol3,
        opacity=\lineopacity,
        dashed,
        line width=\customlinewidthdotted,
    ] table [col sep=tab, y=uncertainty_semdiv_y, x=uncertainty_semdiv_x] {#1};

    \addplot[
        paultol5,
        opacity=\lineopacity,
        line width=\customlinewidth,
    ] table [col sep=tab, y=random_representation_y, x=random_representation_x] {#1};

    \addplot[
        paultol5,
        opacity=\lineopacity,
        dashed,
        line width=\customlinewidthdotted,
    ] table [col sep=tab, y=uncertainty_representation_y, x=uncertainty_representation_x] {#1};
    
    \addplot[
        paultol6,
        opacity=\lineopacity,
        line width=\customlinewidth,
    ] table [col sep=tab, y=random_oracle_y, x=random_oracle_x] {#1};

    \addplot[
        paultol6,
        opacity=\lineopacity,
        line width=\customlinewidthdotted,
        dashed,
    ] table [col sep=tab, y=uncertainty_oracle_y, x=uncertainty_oracle_x] {#1};
  
    \ifstrequal{#5}{passive}%
        {
            \addplot[
                paultol4,
                line width=\custompassivelinewidth,
            ] table [col sep=tab, y=passive_passive_y, x=passive_passive_x] {#1};
        }
        {
            
        }

    \ifstrequal{#6}{legend}{
        \legend{
            $\mathcal{S}_R$$\mathcal{T}_R$,
            $\mathcal{S}_U$$\mathcal{T}_R$,
            $\mathcal{S}_R$$\mathcal{T}_L$,
            $\mathcal{S}_U$$\mathcal{T}_L$,
            $\mathcal{S}_R$$\mathcal{T}_S$,
            $\mathcal{S}_U$$\mathcal{T}_S$,
            $\mathcal{S}_R$$\mathcal{T}_D$,
            $\mathcal{S}_U$$\mathcal{T}_D$,
            $\mathcal{S}_R$$\mathcal{O}$,
            $\mathcal{S}_U$$\mathcal{O}$,
            Passive,
        };
    } {
    
    }

    \end{axis}

}

\usepackage[T1]{fontenc}

\usepackage[utf8]{inputenc}

\usepackage{microtype}
\usepackage{printlen}
\usepackage{booktabs}

\title{Annotator-Centric Active Learning for Subjective NLP Tasks}

\author{
 \textbf{Michiel van der Meer\textsuperscript{1,2}},
 \textbf{Neele Falk\textsuperscript{3}},
 \textbf{Pradeep K. Murukannaiah\textsuperscript{4}},
 \textbf{Enrico Liscio\textsuperscript{4}}
\\
 \textsuperscript{1}Idiap Research Institute, Switzerland\\
 \textsuperscript{2}Leiden Institute of Advanced Computer Science, Leiden University, The Netherlands\\
 \textsuperscript{3}Institute for Natural Language Processing, University of Stuttgart, Germany\\
 \textsuperscript{4}Interactive Intelligence, TU Delft, The Netherlands\\
}

\begin{document}
\maketitle

\begin{abstract}
Active Learning (AL) addresses the high costs of collecting human annotations by strategically annotating the most informative samples. However, for subjective NLP tasks, incorporating a wide range of perspectives in the annotation process is crucial to capture the variability in human judgments. We introduce Annotator-Centric Active Learning (ACAL), which incorporates an annotator selection strategy following data sampling. Our objective is two-fold: \begin{enumerate*}[label=(\arabic*)]
    \item to efficiently approximate the full diversity of human judgments, and
    \item to assess model performance using annotator-centric metrics, which \cameraready{value minority and majority perspectives equally.}
\end{enumerate*}
We experiment with multiple annotator selection strategies across seven subjective NLP tasks, employing both traditional and novel, human-centered evaluation metrics. Our findings indicate that ACAL improves data efficiency and excels in annotator-centric performance evaluations. However, its success depends on the availability of a sufficiently large and diverse pool of annotators to sample from.
\end{abstract}

\section{Introduction}
A challenging aspect of natural language understanding (NLU) is the variability of human judgment and interpretation in subjective tasks (e.g., hate speech detection) \citep{plank2022problem}. In a subjective task, a data sample is typically labeled by a set of annotators, and differences in annotation are reconciled via majority voting, resulting in a single (supposedly, true) ``gold label'' \citep{uma2021learning}. However, this approach has been criticized for treating label variation exclusively as noise, which is especially problematic in sensitive subjective tasks \citep{aroyo2015truth} since it can lead to the exclusion of minority voices \citep{leonardelli2021agreeing}.

Subjectivity can be addressed by modeling the full distribution of annotations for each data sample instead of employing gold labels \citep{plank2022problem}.
However, resources for such approaches are scarce, as most datasets do not (yet) make fine-grained annotation details available \citep{cabitza2023toward}, and representing a full range of perspectives is contingent on obtaining costly annotations from a diverse set of annotators \citep{bakker2022fine}.
\begin{figure}[t]
    \centering
    \includegraphics[width=\columnwidth]{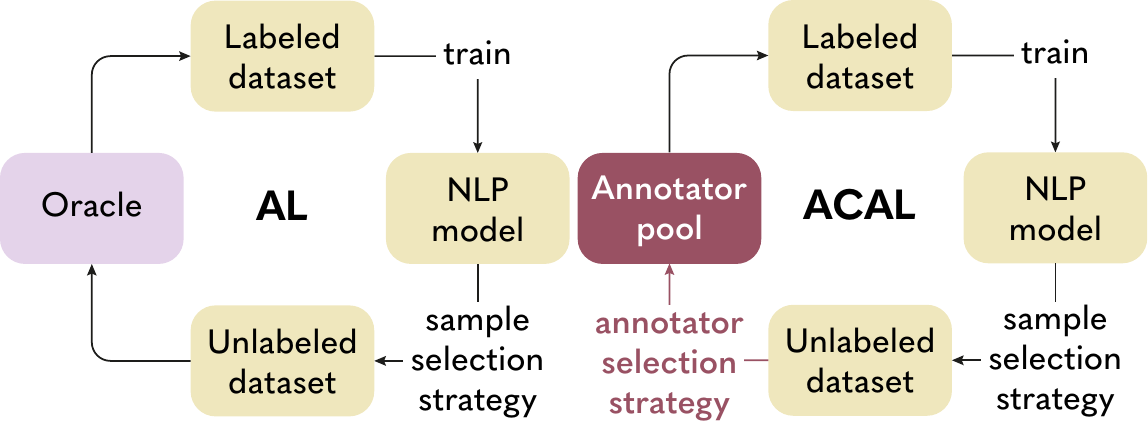}
    \caption{Active Learning (AL) approaches (left) use a sample selection strategy to pick samples to be annotated by an oracle. The Annotator-Centric Active Learning (ACAL) approach (right) extends AL by introducing an annotator selection strategy to choose the annotators who annotate the selected samples.}
    \label{fig:figure1}
\end{figure}

One way to handle a limited annotation budget is to use Active Learning \citep[][AL]{Settles+2012+AL}. Given a pool of unannotated data samples, AL employs a sample selection strategy to obtain maximally informative samples, retrieving the corresponding annotations from a ground truth oracle (e.g., a single human expert).
However, in subjective tasks, there is no such oracle. Instead, we rely on a set of available annotators. Demanding all available annotators to annotate all samples would provide a truthful representation of the annotation distribution, but is often unfeasible, especially if the pool of annotators is large. Thus, deciding \textit{which annotator(s)} should annotate is as critical as deciding which samples to annotate.

In most practical applications, annotators are randomly selected. This results in an annotation distribution insensitive to outlier annotators---most annotations reflect the majority voices and fewer reflect the minority voices. This may not be desirable in applications such as hate speech, where the opinions of the majority and minority should be valued equally. In such cases, a more deliberate annotator selection is required. To ensure a balanced representation of majority and minority voices, we leverage strategies inspired by Rawls' principle of fairness \citep{rawls1973theory}, which advocates that a fair society is achieved when the well-being of the worst-off members of society (the minority annotators, in this case) is maximized.

We introduce Annotator-Centric Active Learning (ACAL) to emphasize and control who annotates which sample. In ACAL (Figure~\ref{fig:figure1}), the sample selection strategy of traditional AL is followed by an \textit{annotator selection strategy}, indicating which of the available annotators should annotate each selected data sample.
\paragraph{Contributions}
\begin{enumerate*}[label=(\arabic*)]
    \item We present ACAL as an extension of the AL approach and introduce three annotator selection strategies aimed at collecting a balanced distribution of minority and majority annotations.
    \item We introduce a suite of annotator-centric evaluation metrics to measure how individual and minority annotators are modeled.
    \item We demonstrate ACAL's effectiveness in three datasets with subjective tasks---hate speech detection, moral value classification, and safety judgments.
\end{enumerate*}

Our experiments show that the proposed ACAL methods can approximate the distribution of human judgments similar to AL while requiring a lower annotation budget and modeling individual and minority voices more accurately. However, our evaluation shows how the task's annotator agreement and the number of available annotations impact ACAL's effectiveness---ACAL is most effective when a large pool of diverse annotators is available. Importantly, our experiments show how the ACAL framework controls how models learn to represent majority and minority annotations. \cameraready{This} is crucial for subjective and sensitive applications \cameraready{such as detecting human values and morality \citep{kiesel:2023a,liscio-etal-2023-text}, argument mining \citep{vandermeer2024empirical}, and hate speech \citep{khurana2024crowd}.}

\section{Related work}
\subsection{Learning with annotator disagreement}
Modeling annotator disagreement is garnering increasing attention \citep{aroyo2015truth,uma2021learning, plank2022problem,cabitza2023toward}. Changing annotation aggregation methods can lead to a fairer representation than simple majority \citep{hovy2013learning, tao2018domain}. Alternatively, the full annotation distribution can be modeled using soft labels \citep{peterson2019human, muller2019does, collins2022eliciting}. Other approaches leverage annotator-specific information, e.g., by including individual classification heads per annotator \citep{davani2022dealing}, embedding annotator behavior \citep{mokhberian2024capturing}, or encoding the annotator's socio-demographic information \citep{beck2024sensitivity}.
\cameraready{Yet, modeling} annotator diversity remains challenging. Standard calibration metrics under human label variation may be unsuitable, especially when the variation is high \citep{Baan2022}. Trade-offs ought to be made between collecting more samples or more annotations \citep{gruber-etal-2024-labels}. Further, solely measuring differences among sociodemographic traits is not sufficient to capture opinion diversity \citep{Orlikowski2023}. Instead, we represent diversity based on \emph{which} annotators annotated \emph{what} and \emph{how}. We experiment with annotator selection strategies to reveal what aspects impact task performance and annotation budget.

\subsection{Active Learning}
\label{related-work:AL}
AL enables a supervised learning model to achieve high performance by judiciously choosing a few training examples \citep{Settles+2012+AL}. In a typical AL scenario, a large collection of unlabeled data is available, and an oracle (e.g., a human expert) is asked to annotate this unlabeled data. A \emph{sampling strategy} is used to iteratively select the next batch of unlabeled data for annotation \citep{Ren2021}. AL has found widespread application in NLP \citep{Zhang2022}. Two main strategies are employed, either by selecting the unlabeled samples on which the model prediction is most uncertain \citep{Zhang2017a}, or by selecting samples that are most representative of the unlabeled dataset \citep{Erdmann2019, Zhao2020}. The combination of AL and annotator diversity is a novel direction. Existing works propose to align model and annotator uncertainties \citep{Baumler2023}, adapt annotator-specific classification heads in AL settings \citep{wang2023actor}, or select texts to annotate based on annotator preferences \citep{kanclerz2023pals}. These methods ignore a crucial part of learning with human variation: the diversity among annotators. We focus on selecting annotators such that they best inform us about the underlying label diversity.

\section{Method}
First, we define the soft-label prediction task we use to train a supervised model. Then, we introduce the traditional AL and the novel ACAL approaches.

\subsection{Soft-label prediction}
Consider a dataset of triples $\{x_i, a_j, y_{ij}\}$, where $x_i$ is a data sample (i.e., a piece of text) and $y_{ij} \in C$ is the class label assigned by annotator $a_j$. The multiple labels assigned to a sample $x_i$ by the different annotators are usually combined into an aggregated label $\hat{y}_i$. For training with soft labels, the aggregation typically takes the form of maximum likelihood estimation \citep{uma2021learning}:
\begin{equation}
    \label{eq:mle-soft-labels}
    \hat{y}_{i}(x) = \frac{\sum^{N}_{i=1}[x_i = x][y_{ij}=c]}{\sum^{N}_{i=1}[x_i = x]}
\end{equation}

In our experiments, we use a passive learning approach that uses all available $\{x_i,\hat{y}_i\}$ to train a model $f_{\theta}$ with cross-entropy loss as a baseline.

\subsection{Active Learning}
AL imposes a sampling technique for inputs $x_i$, such that the most \emph{informative} sample(s) are picked for learning. In a typical AL approach, a set of unlabelled data points $U$ is available. At every iteration, a sample selection strategy $\mathcal{S}$ selects samples $x_i \in U$ to be annotated by an oracle $\mathcal{O}$ that provides the ground truth label distribution $\hat{y}_i$. The selected samples and annotations are added to the labeled data $D$, with which the model $f_{\theta}$ is trained. Alg.~\ref{alg:active-learning} provides an overview of the procedure.

\begin{algorithm}
\SetKwInOut{Input}{input}
\Input{Unlabeled data $U$, Data sampling strategy $\mathcal{S}$, Oracle $\mathcal{O}$}

$D_0 \gets \{\}$

\For{$n = 1 .. N$}
{
    sample data points $x_i$ from $U$ using $\mathcal{S}$\\
    obtain annotation $\hat{y}_i$ for $x_i$ from $\mathcal{O}$
    $D_{n+1} = D_n + \{x_i,\hat{y}_i\}$\\
    train $f_{\theta}$ on $D_{n+1}$\\
}
\caption{AL approach.}
\label{alg:active-learning}
\end{algorithm}
In the sample selection strategies, a batch of data of a given size $B$ is queried at each iteration. Our experiments compare the following strategies:

\noindent\begin{description}[itemsep=-5pt, leftmargin=0pt, topsep=0pt, partopsep=0pt]
    \item[Random ($\mathcal{S}_R$)] selects a $B$ samples uniformly at random from $U$.
    \item[Uncertainty ($\mathcal{S}_U$)] predicts a distribution over class labels with $f_{\theta}(x_i)$ for each $x_i \in U$, and selects $B$ samples with the highest prediction entropy (the samples the model is most uncertain about).
\end{description}

\subsection{Annotator-Centric Active Learning}
\label{sec:method:AAAL}
ACAL builds on AL. In contrast to AL, which retrieves an aggregated annotation $\hat{y}_i$, ACAL employs an annotator selection strategy $\mathcal{T}$ to select one annotator and their annotation for each selected data point $x_i$. Alg.~\ref{alg:annotator-active-learning} describes the ACAL approach.

\begin{algorithm}
\SetKwInOut{Input}{input}
\Input{Unlabeled data $U$, Data sampling strategy $\mathcal{S}$, Annotator sampling strategy $\mathcal{T}$}

$D_0 \gets \{\}$

\For{$n = 1 .. N$}
{
    sample data points $x_i$ from $U$ using $\mathcal{S}$\\
    sample annotators $a_j$ for $x_i$ using $\mathcal{T}$\\
    obtain annotation $y_{ij}$ from $a_j$ for $x_i$\\
    $D_{n+1} = D_n + \{x_i,y_{ij}\}$\\
    train $f_{\theta}$ on $D_{n+1}$\\
}
\caption{ACAL approach.}
\label{alg:annotator-active-learning}
\end{algorithm}

We propose three annotator selection strategies to gather a distribution that uniformly contains all possible (majority and minority) labels, inspired by Rawls' principle of fairness \citep{rawls1973theory}. The strategies vary in the type of information used to represent differences between annotators, including \emph{what} or \emph{how} the annotators have annotated thus far. Our experiments compare the following strategies:
\noindent\begin{description}[itemsep=-5pt, leftmargin=0pt, topsep=0pt, partopsep=0pt]
    \item[Random ($\mathcal{T}_R$)] randomly selects an annotator $a_j$.
    \item[Label Minority ($\mathcal{T}_L$)] considers only \cameraready{information on \textit{how} each annotator has annotated so far (i.e., the} labels that they have assigned). The minority label is selected as the class with the smallest annotation count in the available dataset $D_n$ thus far. Given a new sample, $x_i$, $\mathcal{T}_L$ selects the available annotator that has the largest bias toward the minority label compared to the other available annotators, i.e., who has annotated other samples with the minority label the most.
    \item[Semantic Diversity ($\mathcal{T}_S$)] considers only information on \textit{what} each annotator has annotated so far (i.e., the samples that they have annotated). Given a new sample $x_i$ selected through $\mathcal{S}$, $\mathcal{T}_S$ selects the available annotator for whom $x_i$ is semantically the most different from what the annotator has labeled so far. To measure this difference for an annotator $a_j$, we employ a sentence embedding model to measure the cosine distance between the embeddings of $x_i$ and embeddings of all the samples annotated by $a_j$. We then take the average of all semantic similarities. The annotator with the lowest average similarity score is selected.
    \item[Representation Diversity ($\mathcal{T}_D$)] selects the annotator that has the lowest similarity \cameraready{on average} with all other annotators available for that item. We create a representation for each annotator \cameraready{by averaging the embeddings of samples annotated by $a_j$} together with their respective labels, followed by computing the pair-wise cosine similarity between all annotators.
\end{description}

\section{Experimental Setup}
We describe the experimental setup for the comparisons between ACAL strategies. In all our experiments, we employ a TinyBERT model \citep{jiao2019tinybert} to reduce the number of trainable parameters. Appendix~\ref{app:detailed-experimental-setup} includes a detailed overview of the computational setup and hyperparameters. We make the code for the ACAL strategies and evaluation metrics available via GitHub.\footnote{\url{https://github.com/m0re4u/acal-subjective}}

\subsection{Datasets}
We use three datasets which vary in domain, annotation task (in \textit{italics}), annotator count, and annotations per instance.

\noindent The \textbf{DICES Corpus} \citep{aroyo2023dices} is composed of 990 conversations with an LLM where 172 annotators provided judgments on whether a generated response can be deemed safe (3-way judgments: yes, no, unsure). Samples have 73 annotations on average. We perform a multi-class classification \cameraready{of the judgments}.

\noindent The \textbf{MFTC Corpus} \citep{Hoover2020} is composed of 35K tweets that 23 annotators annotated with any of the 10 moral elements from the Moral Foundation Theory \citep{Graham2013}. We select the elements of \emph{loyalty} (lowest annotation count), \emph{care} (average count), and \emph{betrayal} (highest count). Samples have 4 annotations on average. We create three binary classifications to predict the presence of the respective elements. As most tweets were labeled as non-moral (i.e., with no moral element), we balanced the datasets by subsampling the non-moral class.

\noindent The \textbf{MHS Corpus} \citep{sachdeva2022measuring} consists of 50K social media comments on which 8K annotators judged three hate speech aspects---\emph{dehumanize} (low inter-rater agreement), \emph{respect} (medium agreement), and \emph{genocide} (high agreement)---on a 5-point Likert scale. Samples have 3 annotations on average. We perform a multi-class classification with the annotated Likert scores for each task.

The datasets and tasks differ in levels of annotator agreement, measured via entropy of the annotation distribution. DICES and MHS generally have medium entropy scores, whereas the MFTC entropy is highly polarized (divided between samples with very high and very low agreement). Appendix~\ref{app:dataset-details} provides details of the entropy scores.

\subsection{Evaluation metrics}\label{sec:evalmetrics}
The ACAL strategies aim to guide the model to learn a representative distribution of the annotator's perspectives while reducing annotation effort. To this end, we evaluate the model both with a traditional evaluation metric and a metric aimed at comparing predicted and annotated distributions:
\noindent\begin{description}[itemsep=-5pt, leftmargin=0pt, topsep=0pt, partopsep=0pt]
\item[Macro $F_1$-score ($F_1$)] For each sample in the test set, we select the label predicted by the model with the highest confidence, determine the golden label through a majority agreement aggregation, and compute the resulting macro $F_1$-score.
\item[Jensen-Shannon Divergence ($JS$)] The $JS$ measures the divergence between the distribution of label annotation and prediction \citep{nie2020learn}. We report the average $JS$ for the samples in the test set to measure how well the model can represent the annotation distribution.
\end{description}

Further, since ACAL shifts the focus to annotators, we introduce novel annotator-centric evaluation metrics. First, we report the average among annotators. Second, in line with Rawls' principle of fairness, the result for the worst-off annotators:
\noindent\begin{description}[itemsep=-5pt, leftmargin=0pt, topsep=0pt, partopsep=0pt]
\item[Per-annotator $F_1$ ($F_1^a$) and $JS$ ($JS^a$)] We compute the $F_1$ (or $JS$) for each annotator in the test set using their annotations as golden labels (or target distribution), and average it.
\item[Worst per-annotator $F_1$ ($F_1^w$) and $JS$ ($JS^w$)] We compute the $F_1$ (or $JS$) for each annotator in the test set using their annotations as golden labels (or target distribution), and report the average of the lowest 10\% to mitigate noise.
\end{description}

These metrics allow us to measure the trade-offs between modeling the majority agreement, a representative distribution of annotations, and accounting for minority voices.  In the next section, we describe how we obtained the results.

\subsection{Training procedure}
\label{sec:training-procedure}
We test the annotator selection strategies proposed in Section~\ref{sec:method:AAAL} by comparing all combinations of the two sample selection strategies ($\mathcal{S}_R$ and $\mathcal{S}_U$) and the four annotator selection strategies ($\mathcal{T}_R$, $\mathcal{T}_L$, $\mathcal{T}_S$, and $\mathcal{T}_D$). At each iteration, we use $\mathcal{S}$ to select $B$ unique samples from the unlabeled data pool $U$. We select $B$ as the smallest between 5\% of the number of available annotations and the number of unique samples in the training set. For each selected sample $x_i$, we use $\mathcal{T}$ to select one annotator and retrieve their annotation $y_{ij}$.

We split each dataset into 80\% train, 10\% validation, and 10\% test. We start the training procedure with a warmup iteration of $B$ randomly selected annotations \citep{Zhang2022}. We proceed with the ACAL iterations by combining $\mathcal{S}$ and $\mathcal{T}$. We select the model checkpoint across all AL iterations that led to the best $JS$ performance on the validation set and evaluate it on the test set. We repeat this process across three data splits and model initializations. We report the average scores on the test set.

We compare ACAL with traditional oracle-based AL approaches ($\mathcal{S}_R\mathcal{O}$ and $\mathcal{S}_U\mathcal{O}$), which use the data sampling strategies but obtain all possible annotations for each sample as in Alg.~\ref{alg:active-learning}. Further, we employ a passive learning (PL) approach as an upper bound by training the model on the full dataset, thus observing all available samples and annotations. Similar to ACAL, the AL and PL baselines are averaged over three seeds.

\section{Results}
We start by highlighting the benefits of ACAL over AL and PL (Section~\ref{sec:highlights}). Next, we closely examine ACAL on efficiency and fairness (Section~\ref{sec:eff-fair}). Then, we select a few cases of interest and dive deeper into the strategies' behavior during training (Section~\ref{sec:training-iterations}). Finally, we investigate ACAL across varying levels of subjectivity (Section~\ref{sec:effects-subjectivity}).

\begin{figure}[ht]
    \centering
    \begin{tikzpicture}
    \begin{axis}[
        name={dices-js},
        at={(0,0)},
        anchor={north},
        label style={font=\large},
        tick label style = {font=\large},
        title = {\textbf{\large DICES}},
        xmin=0,
        try min ticks=3,
        xlabel = {\# unique annotations},
        ylabel = {$JS$},
        height=4.5cm,
        width=8cm,
        xmax={40000},
        legend style={
            at={(3.2cm,-1.3cm)},
            anchor=north,
            font={\large},
            /tikz/every even column/.append style={column sep=0.4cm},
            draw=none,
        },
        legend columns=4,
        legend cell align=left,
    ]
    \addplot[
        paultol1,
        opacity=0.8,
        line width=1.5pt,
    ] table [col sep=tab, y=random_active_y, x=random_active_x] {DICES_overall_eval_jsdiv.dat};

    \addplot[
        paultol1,
        dashed,
        opacity=0.8,
        line width=3pt,
    ] table [col sep=tab, y=uncertainty_active_y, x=uncertainty_active_x] {DICES_overall_eval_jsdiv.dat};

    \addplot[
        paultol2,
        opacity=0.8,
        line width=1.5pt,
    ] table [col sep=tab, y=random_lmin_y, x=random_lmin_x] {DICES_overall_eval_jsdiv.dat};

    \addplot[
        paultol2,
        opacity=0.8,
        dashed,
        line width=3pt,
    ] table [col sep=tab, y=uncertainty_lmin_y, x=uncertainty_lmin_x] {DICES_overall_eval_jsdiv.dat};

    \addplot[
        paultol3,
        opacity=0.8,
        line width=1.5pt,
    ] table [col sep=tab, y=random_semdiv_y, x=random_semdiv_x] {DICES_overall_eval_jsdiv.dat};

    \addplot[
        paultol3,
        opacity=0.8,
        dashed,
        line width=3pt,
    ] table [col sep=tab, y=uncertainty_semdiv_y, x=uncertainty_semdiv_x] {DICES_overall_eval_jsdiv.dat};

    \addplot[
        paultol5,
        opacity=0.8,
        line width=1.5pt,
    ] table [col sep=tab, y=random_representation_y, x=random_representation_x] {DICES_overall_eval_jsdiv.dat};

    \addplot[
        paultol5,
        opacity=0.8,
        dashed,
        line width=3pt,
    ] table [col sep=tab, y=uncertainty_representation_y, x=uncertainty_representation_x] {DICES_overall_eval_jsdiv.dat};

    \addplot[
        paultol6,
        opacity=0.8,
        line width=1.5pt,
    ] table [col sep=tab, y=random_oracle_y, x=random_oracle_x] {DICES_overall_eval_jsdiv.dat};

    \addplot[
        paultol6,
        opacity=0.8,
        line width=3pt,
        dashed,
    ] table [col sep=tab, y=uncertainty_oracle_y, x=uncertainty_oracle_x] {DICES_overall_eval_jsdiv.dat};

        \addplot[
            paultol4,
            line width=0.8pt,
        ] table [col sep=tab, y=passive_passive_y, x=passive_passive_x] {DICES_overall_eval_jsdiv.dat};

        \legend{
            $\mathcal{S}_R$$\mathcal{T}_R$,
            $\mathcal{S}_U$$\mathcal{T}_R$,
            $\mathcal{S}_R$$\mathcal{T}_L$,
            $\mathcal{S}_U$$\mathcal{T}_L$,
            $\mathcal{S}_R$$\mathcal{T}_S$,
            $\mathcal{S}_U$$\mathcal{T}_S$,
            $\mathcal{S}_R$$\mathcal{T}_D$,
            $\mathcal{S}_U$$\mathcal{T}_D$,
            $\mathcal{S}_R$$\mathcal{O}$,
            $\mathcal{S}_U$$\mathcal{O}$,
            Passive,
        };

    \end{axis}

    \begin{axis}[
        name={dices-js},
        at={($(dices-js.south)+(0, -3.5cm)$)},
        anchor={north},
        title = {\textbf{\large MHS (dehumanize)}},
        xmin=0,
        label style={font=\large},
        tick label style = {font=\large},
        try min ticks=3,
        xlabel = {\# unique annotations},
        ylabel = {$F_1$},
        height=4.5cm,
        width=8cm,
        xmax={46000},
        legend style={
            at={(3.2cm,-1.3cm)},
            anchor=north,
            /tikz/every even column/.append style={column sep=0.5cm},
            draw=none,
        },
        legend columns=4,
        legend cell align=left,
    ]
    \addplot[
        paultol1,
        opacity=0.8,
        line width=1.5pt,
    ] table [col sep=tab, y=random_active_y, x=random_active_x] {MHS_dehumanize_eval_macro_avg.dat};

    \addplot[
        paultol1,
        dashed,
        opacity=0.8,
        line width=3pt,
    ] table [col sep=tab, y=uncertainty_active_y, x=uncertainty_active_x] {MHS_dehumanize_eval_macro_avg.dat};

    \addplot[
        paultol2,
        opacity=0.8,
        line width=1.5pt,
    ] table [col sep=tab, y=random_lmin_y, x=random_lmin_x] {MHS_dehumanize_eval_macro_avg.dat};

    \addplot[
        paultol2,
        opacity=0.8,
        dashed,
        line width=3pt,
    ] table [col sep=tab, y=uncertainty_lmin_y, x=uncertainty_lmin_x] {MHS_dehumanize_eval_macro_avg.dat};

    \addplot[
        paultol3,
        opacity=0.8,
        line width=1.5pt,
    ] table [col sep=tab, y=random_semdiv_y, x=random_semdiv_x] {MHS_dehumanize_eval_macro_avg.dat};

    \addplot[
        paultol3,
        opacity=0.8,
        dashed,
        line width=3pt,
    ] table [col sep=tab, y=uncertainty_semdiv_y, x=uncertainty_semdiv_x] {MHS_dehumanize_eval_macro_avg.dat};

    \addplot[
        paultol5,
        opacity=0.8,
        line width=1.5pt,
    ] table [col sep=tab, y=random_representation_y, x=random_representation_x] {MHS_dehumanize_eval_macro_avg.dat};

    \addplot[
        paultol5,
        opacity=0.8,
        dashed,
        line width=3pt,
    ] table [col sep=tab, y=uncertainty_representation_y, x=uncertainty_representation_x] {MHS_dehumanize_eval_macro_avg.dat};

    \addplot[
        paultol6,
        opacity=0.8,
        line width=1.5pt,
    ] table [col sep=tab, y=random_oracle_y, x=random_oracle_x] {MHS_dehumanize_eval_macro_avg.dat};

    \addplot[
        paultol6,
        opacity=0.8,
        line width=3pt,
        dashed,
    ] table [col sep=tab, y=uncertainty_oracle_y, x=uncertainty_oracle_x] {MHS_dehumanize_eval_macro_avg.dat};

        \addplot[
            paultol4,
            line width=0.8pt,
        ] table [col sep=tab, y=passive_passive_y, x=passive_passive_x] {MHS_dehumanize_eval_macro_avg.dat};

    \end{axis}
\end{tikzpicture}
    \caption{Learning curves showing model performance on the validation set. On DICES (upper), ACAL approaches are quicker than AL in obtaining similar performance to passive learning. On MHS (lower), ACAL surpasses passive learning in $F_1$ when data has high disagreement.}
    \label{fig:snapshot}
\end{figure}
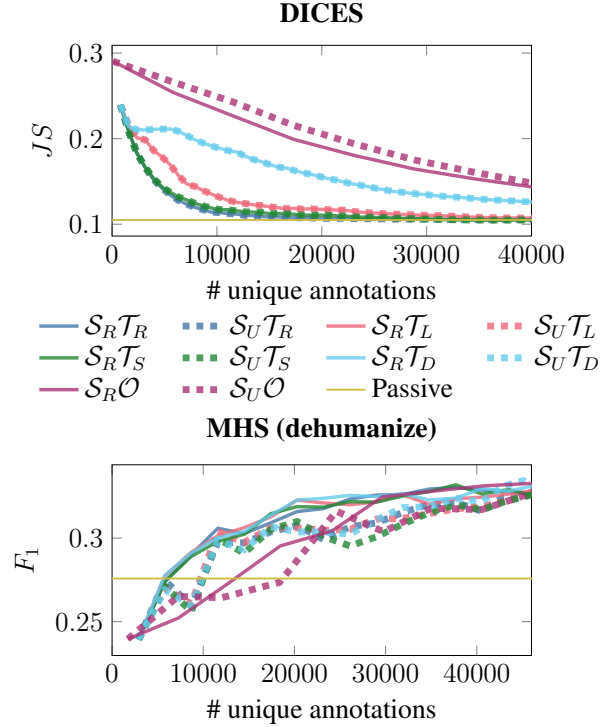

\subsection{Highlights}
\label{sec:highlights}
Our experiments show that ACAL can have a beneficial impact over using PL and AL. Figure~\ref{fig:snapshot} highlights two main findings:
\begin{enumerate*}[label=(\arabic*)]
    \item ACAL strategies can more quickly learn to represent the annotation distribution with a large pool of annotators, and
    \item when agreement between annotators is polarized, ACAL leads to improved results compared to learning from aggregated labels.
\end{enumerate*}
In the next sections, we provide a deeper understanding of the conditions in which ACAL works well.

\subsection{Efficiency and Fairness}
\label{sec:eff-fair}
Table~\ref{tab:annotator-centric-all} presents the results of evaluating the best models (those with the highest $JS$ scores on the validation set) on the test set. We analyze the results along two dimensions:
(a) \textit{efficiency}: what is the impact of the different strategies on the trade-off between annotation budget and performance?
(b) \textit{fairness}: do the selection strategies that aim for a balanced consideration of minority and majority views lead to better performance in the human-centric evaluation metrics?
For MFTC we focus on \textit{care} because it has an average number of samples available, and for MHS we focus on \textit{dehumanize} because it has high levels of disagreement. Appendix~\ref{app:detailed-results} presents \cameraready{the remainder of the} results.

\begin{table}[!ht]
\small
    \centering
    \begin{tabular}{@{}cl@{\hspace{.2cm}}c@{\hspace{.2cm}}c@{\hspace{.2cm}}c@{\hspace{.2cm}}c@{\hspace{.2cm}}c@{\hspace{.2cm}}c@{\hspace{.2cm}}c@{}}
         \toprule
         & & & & \multicolumn{2}{c}{\textbf{Average}} & \multicolumn{2}{c}{\textbf{Worst-off}}\\
         & \textbf{App.} & $F_1$ & $JS$ & $F_1^a $ &  $JS^a$ & $F_1^w$ & $JS^w$ & $\Delta\%$\\
         \midrule
\parbox[t]{2mm}{\multirow{11}{*}{\rotatebox[origin=c]{90}{DICES}}} & $\mathcal{S}_R$$\mathcal{T}_R$      & 53.2 & .100 & 43.2 & \textbf{.186} & 16.7 & .453 & -36.8 \\
& $\mathcal{S}_R$$\mathcal{T}_L$      & 55.5 & .101 & 42.4 & .187 & 15.5 & .450 & -32.7 \\
& $\mathcal{S}_R$$\mathcal{T}_S$      & 61.0 & .103 & \textbf{44.2} & \textbf{.186} & 16.4 & .447 & -35.5 \\
& $\mathcal{S}_R$$\mathcal{T}_D$      & 58.9 & .142 & 43.1 & .203 & 16.9 & \textbf{.370} & -30.0 \\
& $\mathcal{S}_U$$\mathcal{T}_R$      & 53.2 & .100 & 43.2 & \textbf{.186} & 16.7 & .453 & -36.8 \\
& $\mathcal{S}_U$$\mathcal{T}_L$      & 55.5 & .101 & 42.4 & .187 & 15.5 & .450 & -32.7 \\
& $\mathcal{S}_U$$\mathcal{T}_S$      & \textbf{63.1} & \textbf{.098} & 43.9 & .187 & \textbf{18.4} & .447 & \textbf{-38.2} \\
& $\mathcal{S}_U$$\mathcal{T}_D$      & 58.9 & .142 & 43.1 & .203 & 16.9 & \textbf{.370} & -30.0 \\
\cmidrule(ll){2-9}
& $\mathcal{S}_R$$\mathcal{O}$        & 59.1 & .112 & 41.4 & .191 & 13.3 & .425 & -0.1 \\
& $\mathcal{S}_U$$\mathcal{O}$        & 46.2 & .110 & 38.4 & .192 & 11.7 & .427 & -0.1 \\
& PL                                  & 59.0 & .105 & 37.1 & .211 & 12.3 & .479 & -- \\
\midrule
\parbox[t]{2mm}{\multirow{11}{*}{\rotatebox[origin=c]{90}{MFTC (\textit{care})}}} & $\mathcal{S}_R$$\mathcal{T}_R$      & 78.9 & .038 & 61.1 & \textbf{.141} & 37.7 & .247 & -1.6 \\
& $\mathcal{S}_R$$\mathcal{T}_L$      & 78.5 & .037 & \textbf{61.6} & .142 & 39.2 & .249 & -0.4 \\
& $\mathcal{S}_R$$\mathcal{T}_S$      & 78.1 & .039 & 60.0 & .145 & 35.1 & .248 & -1.7 \\
& $\mathcal{S}_R$$\mathcal{T}_D$      & 76.6 & .040 & 60.4 & .144 & 35.7 & \textbf{.243} & -1.7 \\
& $\mathcal{S}_U$$\mathcal{T}_R$      & 79.4 & .038 & 61.2 & .143 & 37.7 & .252 & -5.6 \\
& $\mathcal{S}_U$$\mathcal{T}_L$      & 80.7 &.037 & 58.9 & .142 & \textbf{42.3} & .248 & -2.5 \\
& $\mathcal{S}_U$$\mathcal{T}_S$      & 79.1 & .037 & 60.8 & .143 & 39.9 & .258 & -1.1 \\
& $\mathcal{S}_U$$\mathcal{T}_D$      & 78.1 & .040 & 58.6 & .145 & 35.7 & .253 & -2.5 \\
\cmidrule(ll){2-9}
& $\mathcal{S}_R$$\mathcal{O}$        & 79.0 & .037 & 58.6 & \textbf{.141} & 39.2 & .255 & -0.2 \\
& $\mathcal{S}_U$$\mathcal{O}$        & 79.4 & .037 & 58.3 & .144 & 35.7 & .253 & \textbf{-12.7} \\
& PL             & \textbf{81.1} & \textbf{.032} & 51.2 & .179 & 37.7 & .251 & -- \\
\midrule
\parbox[t]{2mm}{\multirow{11}{*}{\rotatebox[origin=c]{90}{MHS (\textit{dehumanize})}}} & $\mathcal{S}_R$$\mathcal{T}_R$      & \textbf{33.6} & .081 & 31.5 & .394 & 0.0 & .489 & -50.0 \\
& $\mathcal{S}_R$$\mathcal{T}_L$      & 33.1 & .081 & 32.2 & .397 & 0.0 & \textbf{.478} & \textbf{-62.5} \\
& $\mathcal{S}_R$$\mathcal{T}_S$      & 30.5 & .079 & 31.3 & .397 & 0.0 & .480 & \textbf{-62.5} \\
& $\mathcal{S}_R$$\mathcal{T}_D$      & 32.4 & .081 & 31.8 & .398 & 0.0 & .479 & \textbf{-62.5} \\
& $\mathcal{S}_U$$\mathcal{T}_R$      & 32.4 & .080 & 32.2 & .389 & 0.0 & .508 & -7.8 \\
& $\mathcal{S}_U$$\mathcal{T}_L$      & 33.1 & .080 & 32.8 & .388 & 0.0 & .507 & -7.8 \\
& $\mathcal{S}_U$$\mathcal{T}_S$      & \textbf{33.6} & .080 & 32.6 & .388 & 0.0 & .506 & -7.8 \\
& $\mathcal{S}_U$$\mathcal{T}_D$      & 33.0 & .079 & 32.6 & \textbf{.384} & 0.0 & .513 & -3.0 \\
\cmidrule(ll){2-9}
& $\mathcal{S}_R$$\mathcal{O}$        & 32.8 & .077 & \textbf{33.9} & .387 & 0.0 & .496 & -60.1 \\
& $\mathcal{S}_U$$\mathcal{O}$        & 33.3 & .080 & 33.1 & .390 & 0.0 & .497 & -24.7 \\
& PL                                  & 28.0 & \textbf{.075} & 20.2 & .424 & 0.0 & .547 & -- \\

\bottomrule
    \end{tabular}
    \caption{Test set results on the DICES, MFTC (\textit{care}), and MHS (\textit{dehumanize}) datasets. \cameraready{Results report the average test scores from the best-performing model checkpoint on the validation set (lowest $JS$), evaluated across three data splits and model initializations.} $\Delta\%$ denotes the reduction in the annotation budget with respect to passive learning. In bold, the best performance per column and per dataset (higher $F_1$ are better, lower $JS$ are better).}
    \label{tab:annotator-centric-all}
\end{table}

\paragraph{Efficiency}
We discuss the performance on $F_1$ and $JS$ to measure how well the proposed strategies model label distributions and examine the used annotator budget. Across all tasks and datasets, ACAL and AL consistently yield comparable or superior $F_1$ and $JS$ with a lower annotation budget than PL. When comparing ACAL with AL, the results vary depending on the task and dataset. For DICES, there is a significant benefit to using ACAL, as it can save up to $\sim$40\% of the annotation budget while yielding better scores across all metrics than AL. With AL, we observe only a small reduction in annotation cost. For MFTC, AL with $\mathcal{S}_U$ leads to the largest cost benefits ($\sim$12\% less annotation budget), but at a cost in terms of absolute $JS$ and $F_1$. ACAL slightly outperforms AL but does not lead to a decrease in annotation budget. For MHS, both AL and ACAL significantly reduce the annotation cost ($\sim$60\%) while yielding better scores than PL---however, AL and ACAL do not show substantial performance differences. Overall, \cameraready{when looking at $F_1$ and $JS$ which are aggregated over the whole test set,} we conclude that ACAL is most efficient when the pool of available annotators for one sample is large (as with the DICES dataset), whereas the difference between ACAL and AL is negligible with a small pool of annotators per data sample (as with MFTC and MHS).

\paragraph{Fairness}
We investigate the extent to which the models represent individual annotators fairly and capture minority opinions via the annotator-centric evaluation metrics ($F_1^a$, $JS^a$, $F_1^w$, and $JS_w$). We observe a substantial improvement when using AL or ACAL over PL. Further, we observe no single winner-takes-all approach: high $F_1$ and $JS$ scores do not consistently co-occur with high scores for the annotator-centric metrics. This highlights the need for a more comprehensive evaluation to assess models for subjective tasks. \cameraready{Yet, w}e observe that ACAL slightly outperforms AL in modeling individual annotators ($JS^a$ and $F_1^a$). This trend is particularly evident with DICES, again likely due to the large pool of annotators available per data sample. Lastly, ACAL is best in the worst-off metrics ($JS^w$ and $F_1^w$), showing the ability to better represent minority opinions as a direct consequence of the proposed annotator selection strategies on DICES and MFTC. However, all approaches score 0 for $F_1^w$ on MHS. This is due to the high disagreement in this dataset: the 10\% worst-off annotators always disagree with a hard label derived from the predicted label distribution. In conclusion, our experiments show that, when a large pool of annotators is available, a targeted sampling of annotators requires fewer annotations and is fairer. That is, minority opinions are better represented without large sacrifices in performance compared to the overall label distribution.

\begin{figure*}[th]
    \centering
    \includestandalone[width=\textwidth]{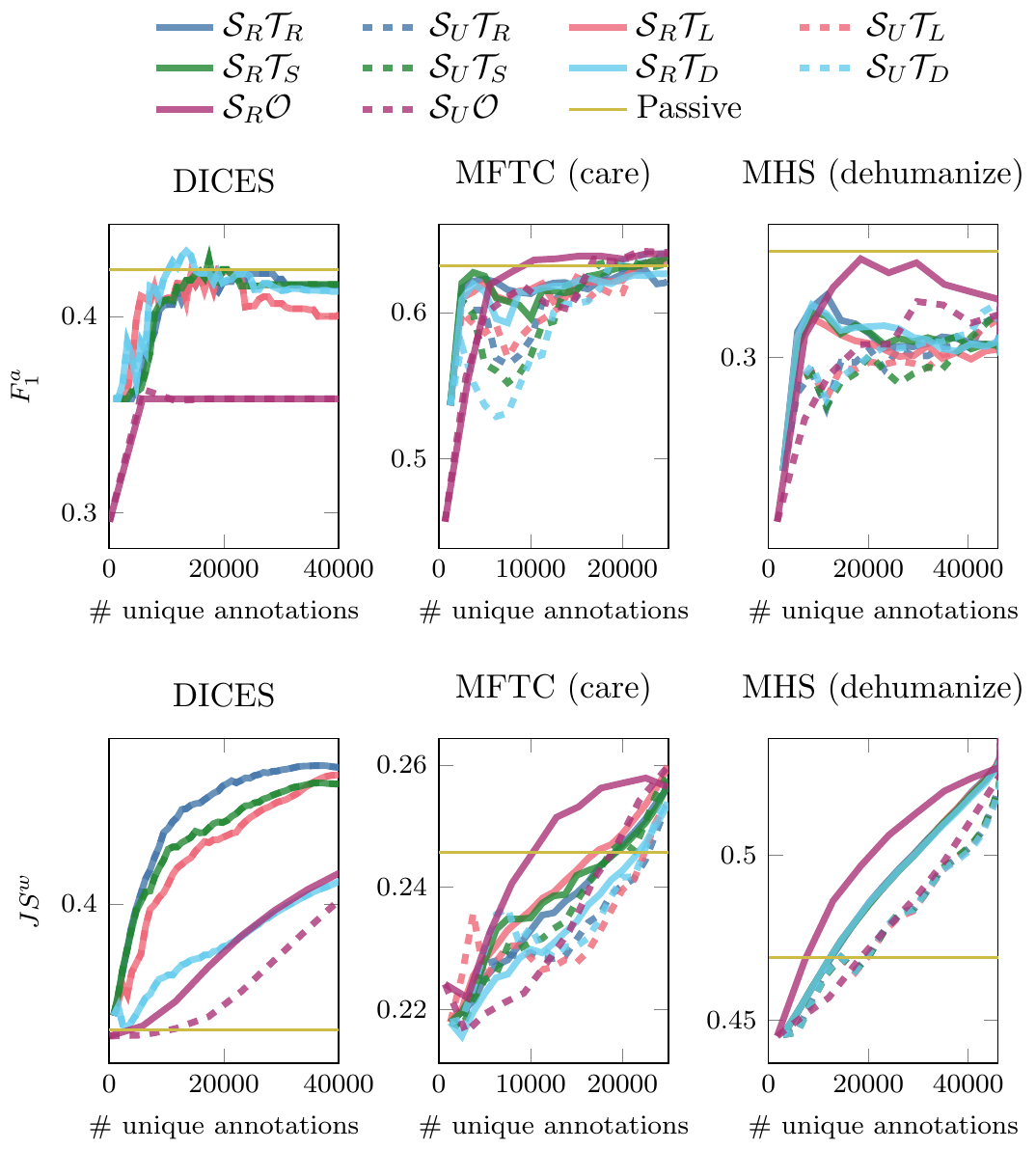}
    \caption{Selected plots showing the $F_1^a$ and $JS^w$ performance on the validation set during the ACAL and AL iterations for DICES, MFTC (\textit{care}), and MHS (\textit{dehumanize}). Higher $F_1^a$ is better, lower $JS^w$ is better. Y-axes are scaled to highlight the relative performance to PL.}
    \label{fig:selected-training-plots}
\end{figure*}

\subsection{Convergence}
\label{sec:training-iterations}
The evaluation on the test set paints a general picture of the advantage of using ACAL over AL or PL. In this section, we assess how different ACAL strategies converge over iterations.
We describe the major patterns across our experiments by analyzing six examples of interest with $F_1^a$ and $JS^w$ (Figure~\ref{fig:selected-training-plots}). We select $F_1^a$ because it reveals how well individual annotators are modeled on average, and  $JS^w$ to measure how strategies deviate from modeling the majority perspective.  Appendix~\ref{app:detailed-results-training} provides an overview of all metrics.

First, we notice that the trends for $F_1^a$ and $JS^w$ are both increasing---the first is expected, but the second requires an explanation. As the model is exposed to more annotations over the training iterations, the predicted label distribution starts to fit the true label distribution. However, here we consider each annotator individually: $JS^w$ reports the average of the 10\% lowest $JS$ scores per annotator. The presence of disagreement implies the existence of annotators that annotate differently from the majority. Since our models predict the full distribution, they assign a proportional probability to dissenting annotators. Thus, learning to model the full distribution of annotations leads to an increase in $JS^w$.

Second, we notice a difference between ACAL and AL. On MFTC and MHS, ACAL, compared to AL, yields overall smaller $JS^w$ at the cost of a slower convergence in $F_1^a$, showing the trade-off between modeling all annotators and representing minorities. However, with DICES the trend is the opposite. This is due to AL having access to the complete label distribution: it can model a balanced distribution, leading to lower worst-off performance. With a large number of annotations, ACAL requires more iterations to get the same balanced predicted distribution.

Third, we observe differences among the annotator selection strategies ($\mathcal{T}$). $\mathcal{T}_D$ shows the most differences---both $JS^w$ and $F_1^a$ increase slower than for the other strategies. This suggests that selecting annotators based on the average embedding of the annotated content strongest emphasizes diverging label behavior.

Finally, we analyze the impact of the sample selection strategies ($\mathcal{S}$, dotted vs. solid lines in Figure~\ref{fig:selected-training-plots}). For DICES, $\mathcal{S}_R$ and $\mathcal{S}_U$ lead to comparable results, likely due to the low number of samples.
Using $\mathcal{S}_U$ in MFTC leads to $F_1^a$ performance decreasing at the start of training. The strategy prioritizes obtaining annotations for already added samples to lower their entropy, while the variation in labels is irreconcilable (since there are limited labels available, and they are in disagreement). We see a similar pattern for MHS.

These results further underline our main finding that ACAL is effective in representing diverse annotation perspectives when there is a \begin{enumerate*}[label=(\arabic*)]
    \item heterogeneous pool of annotators, and
    \item a task that facilitates human label variation.
\end{enumerate*}

\subsection{Impact of subjectivity}
\label{sec:effects-subjectivity}
We further investigate ACAL strategies on \begin{enumerate*}[label=(\arabic*)]
    \item label entropy, and
    \item cross-task performance.
\end{enumerate*}

\paragraph{Alignment of ACAL strategies during training}
We want to investigate how well the ACAL strategies align with the overall subjective annotations: do they drive the model entropy in the right direction? We measure the entropy of the samples in the labeled training set at each iteration and compare it to the entropy of all annotations of those samples. Higher entropy \cameraready{in the labeled training set than the actual entropy} suggests that the selection strategy overestimates uncertainty. Lower entropy indicates that the model may not sufficiently account for disagreement. When the entropy matches the true entropy, the selection strategy is well-calibrated \cameraready{to strike a healthy middle ground between sampling diverse labels and finding the majority class}. We focus on DICES as a case study due to the wide range of entropy scores. We group each sample based on the true label entropy into low ($<0.43$), medium ($0.43-0.72$), and high ($> 0.72$). We apply the same categorization at each training iteration for samples labeled thus far. Subsequently, we plot the proportion of data points for which the selection strategy results in excessively high or excessively low entropy.

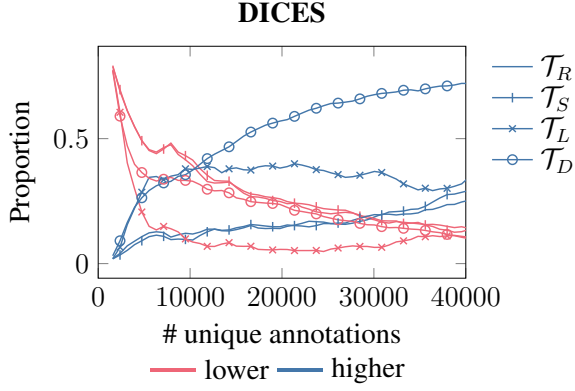
\begin{figure}[t]
    \centering
    \begin{tikzpicture}[
legendcolornode/.style={
    rectangle,
    fill=green,
    minimum height=1.6pt,
    inner sep=0,
    minimum width=6.66mm,
    },
]
    \begin{axis}[
        name={active},
        at={(0,0)},
        anchor={north},
        label style={font=\large},
        tick label style = {font=\large},
        title = {\textbf{\large DICES}},
        xmin=0,
        try min ticks=3,
        xlabel = {\# unique annotations},
        ylabel = {Proportion},
        height=5cm,
        width=7cm,
        xmax={40000},
        legend style={
            font={\large},
            /tikz/every even column/.append style={column sep=0.4cm},
            draw=none,
        },
        legend columns=4,
        legend cell align=left,
    ]

    \addplot[
        paultol2,
        opacity=1.0,
        line width=0.6pt,
    ] table [col sep=tab, y=lower than H, x=total_unique_annotations_seen] {active_uncertainty_entropy.csv};
    \label{p1}

    \addplot[
        paultol1,
        opacity=1.0,
        line width=0.6pt,
    ] table [col sep=tab, y=higher than H, x=total_unique_annotations_seen] {active_uncertainty_entropy.csv};
    \label{p3}

    \addplot[
        paultol2,
        opacity=1.0,
        line width=0.6pt,
        mark=|,
        mark repeat=3,mark phase=2,
    ] table [col sep=tab, y=lower than H, x=total_unique_annotations_seen] {semdiv_uncertainty_entropy.csv};

    \addplot[
        paultol1,
        opacity=1.0,
        line width=0.6pt,
        mark=|,
        mark repeat=3,mark phase=2,
    ] table [col sep=tab, y=higher than H, x=total_unique_annotations_seen] {semdiv_uncertainty_entropy.csv};
    \label{p4}

    \addplot[
        paultol2,
        opacity=1.0,
        line width=0.6pt,
        mark=x,
        mark repeat=3,mark phase=2,
    ] table [col sep=tab, y=lower than H, x=total_unique_annotations_seen] {lmin_uncertainty_entropy.csv};

    \addplot[
        paultol1,
        opacity=1.0,
        line width=0.6pt,
        mark=x,
        mark repeat=3,mark phase=2,
    ] table [col sep=tab, y=higher than H, x=total_unique_annotations_seen] {lmin_uncertainty_entropy.csv};
    \label{p5}

    \addplot[
        paultol2,
        opacity=1.0,
        line width=0.6pt,
        mark=o,
        mark repeat=3,mark phase=2,
    ] table [col sep=tab, y=lower than H, x=total_unique_annotations_seen] {representation_uncertainty_entropy.csv};

    \addplot[
        paultol1,
        opacity=1.0,
        line width=0.6pt,
        mark=o,
        mark repeat=3,mark phase=2,
    ] table [col sep=tab, y=higher than H, x=total_unique_annotations_seen] {representation_uncertainty_entropy.csv};
    \label{p6}
    \end{axis}

    \node[
        matrix,
        font=\large,
    ] at (0cm,-4.77cm) {
        \node[legendcolornode,fill=paultol2]  {}; & \node {lower}; & \node[legendcolornode,fill=paultol1] {}; & \node {higher}; \\
    };


    \node[
        font=\large
    ] at (3.7,-1) {\shortstack[l]{
        \ref{p3} $\mathcal{T}_R$ \\
        \ref{p4} $\mathcal{T}_S$ \\
        \ref{p5} $\mathcal{T}_L$ \\
        \ref{p6} $\mathcal{T}_D$ }
    };

\end{tikzpicture}
    \caption{Proportion of data samples that result in higher or lower entropy than the target label distribution per ACAL strategy.}
    \label{fig:align-training}
\end{figure}

Figure~\ref{fig:align-training} visualizes the proportions. At the beginning of training, entropy is generally low because samples have few annotations. Over time, the selected annotations better align with the true entropy. \cameraready{At the start (at 10K unique annotations), roughly only a third of the samples have aligned entropy scores
($T_R=27\%,T_S=27\%,T_L=33\%,T_D=32\%$). Further towards the end of the ACAL iterations, this has increased for all ACAL strategies except $T_D$ ($T_R=64\%,T_S=62\%,T_L=57\%,T_D=17\%$).}
When and how much the strategies succeed in matching the true label distribution differs: $\mathcal{T}_S$ and $\mathcal{T}_R$ take longer to increase label entropy than the other two strategies. They are conservative in adding diverse labels. $\mathcal{T}_L$ and $\mathcal{T}_D$ increase the proportion of well-aligned data points earlier in the training process, achieving a balanced entropy alignment sooner. However, both strategies start to overshoot the target entropy, whereas the others show a more gradual alignment with the true entropy. This effect is strongest for $\mathcal{T}_D$. This finding suggests that minority-aware annotator-selection \cameraready{($\mathcal{T}_L$ and $\mathcal{T}_D$)} strategies achieve the best results in the early stages of training---that is, they are effective for quickly raising entropy but can lead to overrepresentation.

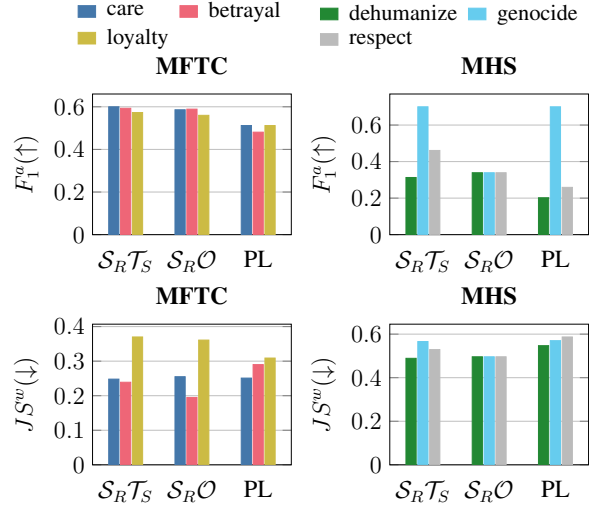
\begin{figure}[t]
    \centering
    \begin{tikzpicture}
    \begin{axis}[
        name={f1avg},
        width  = 5cm,
        height = 4cm,
        major x tick style = transparent,
        ybar=2*\pgflinewidth,
        bar width=5pt,
        label style={font=\large},
        tick label style = {font=\large},
        title={\textbf{\large MFTC}},
        ymajorgrids = true,
        ylabel = {$F_1^a (\uparrow)$},
        symbolic x coords={$\mathcal{S}_R\mathcal{T}_S$,$\mathcal{S}_R\mathcal{O}$, PL},
        xtick = data,
        scaled y ticks = false,
        enlarge x limits=0.25,
        ymin=0,
        legend cell align=left,
        legend columns=2,
        legend style={
            draw=none,
            anchor=south east,
            at={(1,1.27)},
            column sep=0.3em,
        },
        legend image code/.code={
            \draw [#1] (0cm,-0.1cm) rectangle (0.3cm,0.1cm);
        },
    ]
        \addplot[style={paultol1,fill=paultol1}]
            coordinates {($\mathcal{S}_R\mathcal{T}_S$, 0.600) ($\mathcal{S}_R\mathcal{O}$,0.586) (PL,0.512)};

        \addplot[style={paultol2,fill=paultol2}]
             coordinates {($\mathcal{S}_R\mathcal{T}_S$,0.593) ($\mathcal{S}_R\mathcal{O}$,0.589) (PL,0.481)};

        \addplot[style={paultol4,fill=paultol4}]
             coordinates {($\mathcal{S}_R\mathcal{T}_S$,0.573) ($\mathcal{S}_R\mathcal{O}$,0.560) (PL,0.512)};
        \legend{care, betrayal, loyalty}
    \end{axis}

    \begin{axis}[
        at={($(f1avg.south)+(0em, -4em)$)},
        anchor={north},
        width  = 5cm,
        height = 4cm,
        major x tick style = transparent,
        ybar=2*\pgflinewidth,
        label style={font=\large},
        tick label style = {font=\large},
        bar width=5pt,
        title={\textbf{\large MFTC}},
        ymajorgrids = true,
        ylabel = {$JS^w (\downarrow)$},
        symbolic x coords={$\mathcal{S}_R\mathcal{T}_S$,$\mathcal{S}_R\mathcal{O}$, PL},
        xtick = data,
        scaled y ticks = false,
        enlarge x limits=0.25,
        ymin=0,
    ]
        \addplot[style={paultol1,fill=paultol1,mark=none}]
            coordinates {($\mathcal{S}_R\mathcal{T}_S$, 0.248) ($\mathcal{S}_R\mathcal{O}$,0.255) (PL,0.251)};

        \addplot[style={paultol2,fill=paultol2,mark=none}]
             coordinates {($\mathcal{S}_R\mathcal{T}_S$,0.239) ($\mathcal{S}_R\mathcal{O}$,0.195) (PL,0.290)};

        \addplot[style={paultol4,fill=paultol4,mark=none}]
             coordinates {($\mathcal{S}_R\mathcal{T}_S$,0.370) ($\mathcal{S}_R\mathcal{O}$,0.361) (PL,0.309)};
    \end{axis}

    \begin{axis}[
        name={f1avg_mhs},
        at={($(f1avg.east)+(4.4em, 0em)$)},
        anchor=west,
        width  = 5cm,
        height = 4cm,
        major x tick style = transparent,
        ybar=2*\pgflinewidth,
        bar width=5pt,
        title={\textbf{\large MHS}},
        label style={font=\large},
        tick label style = {font=\large},
        ymajorgrids = true,
        ylabel = {$F_1^a (\uparrow)$},
        symbolic x coords={$\mathcal{S}_R\mathcal{T}_S$,$\mathcal{S}_R\mathcal{O}$, PL},
        xtick = data,
        scaled y ticks = false,
        enlarge x limits=0.25,
        ymin=0,
        legend cell align=left,
        legend columns=2,
        legend style={
            draw=none,
            anchor=south east,
            at={(1,1.27)},
            column sep=0.3em,
        },
        legend image code/.code={
            \draw [#1] (0cm,-0.1cm) rectangle (0.3cm,0.1cm);
        },
    ]
        \addplot[style={paultol3,fill=paultol3,mark=none}]
            coordinates {($\mathcal{S}_R\mathcal{T}_S$, 0.313) ($\mathcal{S}_R\mathcal{O}$,0.339) (PL,0.202)};
        \addplot[style={paultol5,fill=paultol5,mark=none}]
             coordinates {($\mathcal{S}_R\mathcal{T}_S$,0.700) ($\mathcal{S}_R\mathcal{O}$,0.339) (PL,0.700)};
        \addplot[style={paultol7,fill=paultol7,mark=none}]
             coordinates {($\mathcal{S}_R\mathcal{T}_S$,0.461) ($\mathcal{S}_R\mathcal{O}$,0.339) (PL,0.259)};
        \legend{dehumanize, genocide, respect}
    \end{axis}

    \begin{axis}[
        at={($(f1avg_mhs.south)+(0, -4em)$)},
        anchor={north},
        width  = 5cm,
        height = 4cm,
        major x tick style = transparent,
        label style={font=\large},
        tick label style = {font=\large},
        ybar=2*\pgflinewidth,
        bar width=5pt,
        title={\textbf{\large MHS}},
        ymajorgrids = true,
        ylabel = {$JS^w (\downarrow)$},
        symbolic x coords={$\mathcal{S}_R\mathcal{T}_S$,$\mathcal{S}_R\mathcal{O}$, PL},
        xtick = data,
        scaled y ticks = false,
        enlarge x limits=0.25,
        ymin=0,
    ]
        \addplot[style={paultol3,fill=paultol3,mark=none}]
            coordinates {($\mathcal{S}_R\mathcal{T}_S$, 0.489) ($\mathcal{S}_R\mathcal{O}$,0.496) (PL,0.547)};

        \addplot[style={paultol5,fill=paultol5,mark=none}]
             coordinates {($\mathcal{S}_R\mathcal{T}_S$,0.566) ($\mathcal{S}_R\mathcal{O}$,0.496) (PL, 0.570)};

        \addplot[style={paultol7,fill=paultol7,mark=none}]
             coordinates {($\mathcal{S}_R\mathcal{T}_S$,0.529) ($\mathcal{S}_R\mathcal{O}$,0.496) (PL,0.587)};
    \end{axis}

\end{tikzpicture}
    \caption{Comparison of ACAL, AL, and PL across different MFTC and MHS tasks. Higher $F_1^a$ is better, and lower $JS^w$ is better.}
    \label{fig:task-comparison}
\end{figure}

\paragraph{Cross-task performance}
Figure~\ref{fig:task-comparison} compares the two annotator-centric metrics on the three tasks of MFTC and MHS---the datasets for which we have seen the least impact of ACAL over AL and PL. We select a data sampling ($\mathcal{S}_R$) and annotator sampling strategy ($\mathcal{T}_S$), based on its strong performance on DICES for comprehensive comparison.

When evaluating MFTC \emph{loyalty}, which has the highest disagreement, $JS^w$ is more accurately approximated with PL. Similarly, ACAL is outperformed by AL on $F_1^a$ for the \emph{dehumanize} (high disagreement) task. However, for the less subjective task \emph{genocide}, ACAL leads to higher $F_1^a$. This suggests that the effectiveness of annotation strategies varies depending on the task's degree of subjectivity \textit{and} the available pool of annotators. The more heterogeneous the annotation behavior, indicative of a highly subjective task, the larger the pool of annotators required for each sample selection. We also observe that there is a trade-off between modeling the majority of annotators equally ($F_1^a$) and prioritizing the minority ($JS^w$).

\section{Conclusion}
We present ACAL as an extension of AL to emphasize the selection of diverse annotators. We introduce three novel annotator selection strategies and four annotator-centric metrics and experiment with tasks across three different datasets. We find that the ACAL approach is especially effective in reducing the annotation budget when the pool of available annotators is large. However, its effectiveness is contingent on data characteristics such as the number of annotations per sample, the number of annotations per annotator, and the nature of disagreement in the task annotations. Furthermore, our novel evaluation metrics display the trade-off between modeling overall distributions of annotations and adequately accounting for minority voices, showing that different strategies can be tailored to meet different goals. Especially early in the training process, strategies that are aggressive in obtaining diverse labels have a beneficial impact \cameraready{in accounting for minority voices}. \cameraready{However}, we recognize that gathering a distribution that uniformly contains all possible (minority and majority) labels can be overly sensitive to small minorities or noise. Future work should integrate methods that account for noisy annotations \citep{webergenzel2024varierr}.
Striking a balance between utilitarian and egalitarian approaches, \cameraready{such as between modeling aggregated distributions and accounting for minority voices \citep{LeraLeri2024} is crucial for inferring context-dependent values \citep{liscio2023valueinf, vandermeer2023differences}}.

\section*{Limitations}
The main limitation of this work is that the experiments are based on simulated AL which is known to bear several shortcomings \citep{margatina2023limitations}. In our study, a primary challenge arises with two of the datasets (MFTC, MHS), which, despite having a large pool of annotators, lack annotations from every annotator for each item. Consequently, in real-world scenarios, the annotator selection strategies for these datasets would benefit from access to a more extensive pool of annotators. This limitation likely contributes to the underperformance of ACAL on these datasets compared to DICES. We emphasize the need for more datasets that feature a greater number of annotations per item, as this would significantly enhance research efforts aimed at modeling human disagreement.

Since we evaluate four different annotator selection strategies and two sample selection strategies across three datasets and seven tasks, the amount of experiments is high. This did not allow for further investigation of other methods for measuring uncertainty such as ensemble methods \citep{lakshminarayanan2017simple}, different classification models, the extensive turning of hyperparameters, or even different training paradigms like low-rank adaptation \citep{hu2022lora}.
Lastly, a limitation of our annotator selection strategies is that they rely on a small annotation history. This is why we require a warmup phase for some of the strategies, for which we decided to take a random sample of annotations. Incorporating informed warmup strategies, incorporating ACAL strategies that do not rely on annotator history, or making use of more elaborate hybrid human--AI approaches \citep{vandermeer2024hybrid} may positively impact its performance and data efficiency.

\section*{Ethical Considerations}
Our goal is to approximate a good representation of human judgments over subjective tasks. We want to highlight the fact that the \textit{performance} of the models differs a lot depending on which metric is used. We tried to account for a less majority-focussed view when evaluating the models which is very important, especially for more human-centered applications, such as hate-speech detection. However, the evaluation metrics we use do not fully capture the diversity of human \emph{judgments}, but just that of \emph{labeling behavior}. The selection of metrics should align with the specific goals and motivations of the application, and there is a pressing need to develop more metrics to accurately reflect human variability in these tasks.

Our experiments are conducted on English datasets due to the scarcity of unaggregated datasets in other languages. In principle, ACAL can be applied to other languages (given the availability of multilingual models to semantically embed textual items for some particular strategies used in this work). We encourage the community to enrich the dataset landscape by incorporating more perspective-oriented datasets in various languages, ACAL potentially offers a more efficient method for creating such datasets in real-world scenarios.

\section*{Acknowledgements}
This research was partially funded by the Netherlands Organisation for Scientific Research (NWO) through the Hybrid Intelligence Centre via the Zwaartekracht grant (024.004.022) and by the Hasler Foundation through the FactCheck project at Idiap. We would like to thank Gabriella Lapesa for her valuable feedback on earlier versions of this paper. We would also like to thank the ARR reviewers for their helpful feedback.

\bibliography{old, new, michiel}
\bibliographystyle{acl_natbib}

\clearpage
\appendix

\renewcommand{\thefigure}{A\arabic{figure}}
\setcounter{figure}{0}
\renewcommand{\thetable}{A\arabic{table}}
\setcounter{table}{0}

\section{Detailed Experimental Setup}
\label{app:detailed-experimental-setup}
\begin{table*}[ht]
    \small
    \centering
    \begin{tabular}{llccccc}
    \toprule
    \textbf{Dataset} & \textbf{Task (\textit{dimension})} & \textbf{\# Samples} & \textbf{\# Annotators} & \textbf{\# Annotations} &\textbf{\# Annotations per item}\\
    \midrule
    DICES  & Safety Judgment & 990 & 172 & 72,103 & 72.83\\
    MFTC   & Morality (\textit{care}) & 8,434 & 23 & 31,310 & 3.71\\
    MFTC   & Morality (\textit{loyalty}) & 3,288 & 23 & 12,803 & 3.89 \\
    MFTC   & Morality (\textit{betrayal}) &  12,546 & 23 & 47,002 & 3.75\\
    MHS    & \makecell[l]{Hate Speech (\textit{dehumanize},\\\textit{genocide}, \textit{respect})} & 17,282 & 7,807 & 57,980 & 3.35\\
    \bottomrule
    \end{tabular}
    \caption{Overview of the datasets and tasks employed in our work.}
    \label{tab:datasets}
\end{table*}

\subsection{Dataset details}
We provide an overview of the datasets used in our work in Table~\ref{tab:datasets}. We split the data on samples, meaning that all annotations for any given sample are completely contained in each separate split.

\subsection{Hyperparameters}
We report the hyperparameters for training passive, AL, and ACAL in Tables~\ref{app:tab:hypers-passive},~\ref{app:tab:hypers-al}, and~\ref{app:tab:hypers}, respectively. For turning the learning rate for passive learning, on each dataset, we started with a learning rate of 1e-06 and increased it by a factor of 3 in steps until the model showed a tendency to overfit quickly (within a single epoch). All other parameters are kept on their default setting.
\label{app:hyperparams}

\begin{table}[ht]
    \centering
    \begin{tabular}{lr}
    \toprule
         \textbf{Parameter} & \textbf{Value}  \\
    \midrule
         learning rate  & 1e-04 (constant) \\
         max epochs & 50\\
         early stopping  & 3 \\
         batch size  & 128\\
         weight decay  & 0.01\\
    \bottomrule
    \end{tabular}
    \caption{Hyperparameters for the passive learning. }
    \label{app:tab:hypers-passive}
\end{table}

\begin{table}[ht]
    \centering
    \begin{tabular}{@{}l@{\hspace{0.1cm}}l@{\hspace{0.1cm}}r@{}}
    \toprule
         \textbf{Parameter} & \textbf{Dataset (task)} & \textbf{Value}  \\
    \midrule
         learning rate & all & 1e-05\\
         batch size & all & 128 \\
         \makecell[l]{epochs per\\ round}& all & 20 \\
         num iterations & all & 10 \\
         sample size & DICES & 79\\
         sample size & MFTC (care) & 674\\
         sample size & MFTC (betrayal) & 1011\\
         sample size & MFTC (loyalty) & 263\\
         sample size & \makecell[l]{MHS (dehumanize), MHS\\ (genocide), MHS (respect)} & 1728\\
    \bottomrule
    \end{tabular}
    \caption{Hyperparameters for the oracle-based active learning approaches. }
    \label{app:tab:hypers-al}
\end{table}

\begin{table}[ht]
    \centering
    \begin{tabular}{lp{3.3cm}r}
    \toprule
         \textbf{Parameter} & \textbf{Dataset} & \textbf{Value}  \\
    \midrule
         learning rate & all & 1e-05 \\
         num iterations & DICES & 50 \\
         num iterations & MFTC (all), MHS (all) & 20 \\
         \makecell[l]{epochs per\\ round}& DICES, MHS (all)& 20 \\
         \makecell[l]{epochs per\\ round}& MFTC (all) & 30 \\
         sample size & DICES & 792 \\
         sample size & MFTC (care) & 1250\\
         sample size & MFTC (betrayal) & 1894\\
         sample size & MFTC (loyalty) & 512\\
         sample size & MHS (dehumanize), MHS (genocide), MHS (respect) & 2899\\
    \bottomrule
    \end{tabular}
    \caption{Hyperparameters for the annotator-centric active learning approaches. }
    \label{app:tab:hypers}
\end{table}

\subsection{Training details}
Experiments were largely run between January and April 2024. Obtaining the ACAL results for a single run takes up to an hour on a Nvidia RTX4070. For large-scale computation, our experiments were run on a cluster with heterogeneous computing infrastructure, including RTX2080 Ti, A100, and Tesla T4 GPUs. Obtaining the results of all experiments required a total of 231 training runs, combining:
\begin{enumerate*}[label=(\arabic*)]
    \item two data sampling strategies,
    \item four annotator sampling strategies, plus an additional Oracle-based AL approach,
    \item a passive learning approach.
\end{enumerate*}
Each of the above were run for \begin{enumerate*}[label=(\arabic*)]
    \item three folds, each with a different seed, and
    \item the seven tasks across three datasets.
\end{enumerate*}
For training all our models, we employ the AdamW optimizer \citep{loshchilov2018decoupled}. Our code is based on the Huggingface library \citep{wolf2019huggingface}, unmodified values are taken from their defaults.

\subsection{ACAL annotator strategy details}
We provide additional information about the implementations of the strategies used for selecting annotators to provide a label to a sample.
\begin{description}[itemsep=-5pt, leftmargin=0pt, topsep=0pt, partopsep=0pt]
    \item[$\mathcal{T}_S$] uses a sentence embedding model to represent the content that an annotator has annotated. We use \texttt{all-MiniLM-L6-v2}\footnote{\url{https://huggingface.co/sentence-transformers/all-MiniLM-L6-v2}}. We select annotators that have not annotated yet (empty history) before picking from those with a history to prioritize filling the annotation history for each annotator.
    \item [$\mathcal{T}_D$] creates an average embedding for the content annotated by each annotator and selects the most different annotator. We use the same sentence embedding model as $\mathcal{T}_S$. To avoid overfitting, we perform PCA and retain only the top 10 most informative principal components for representing each annotator.
\end{description}

\subsection{Disagreement rates}
\label{app:dataset-details}
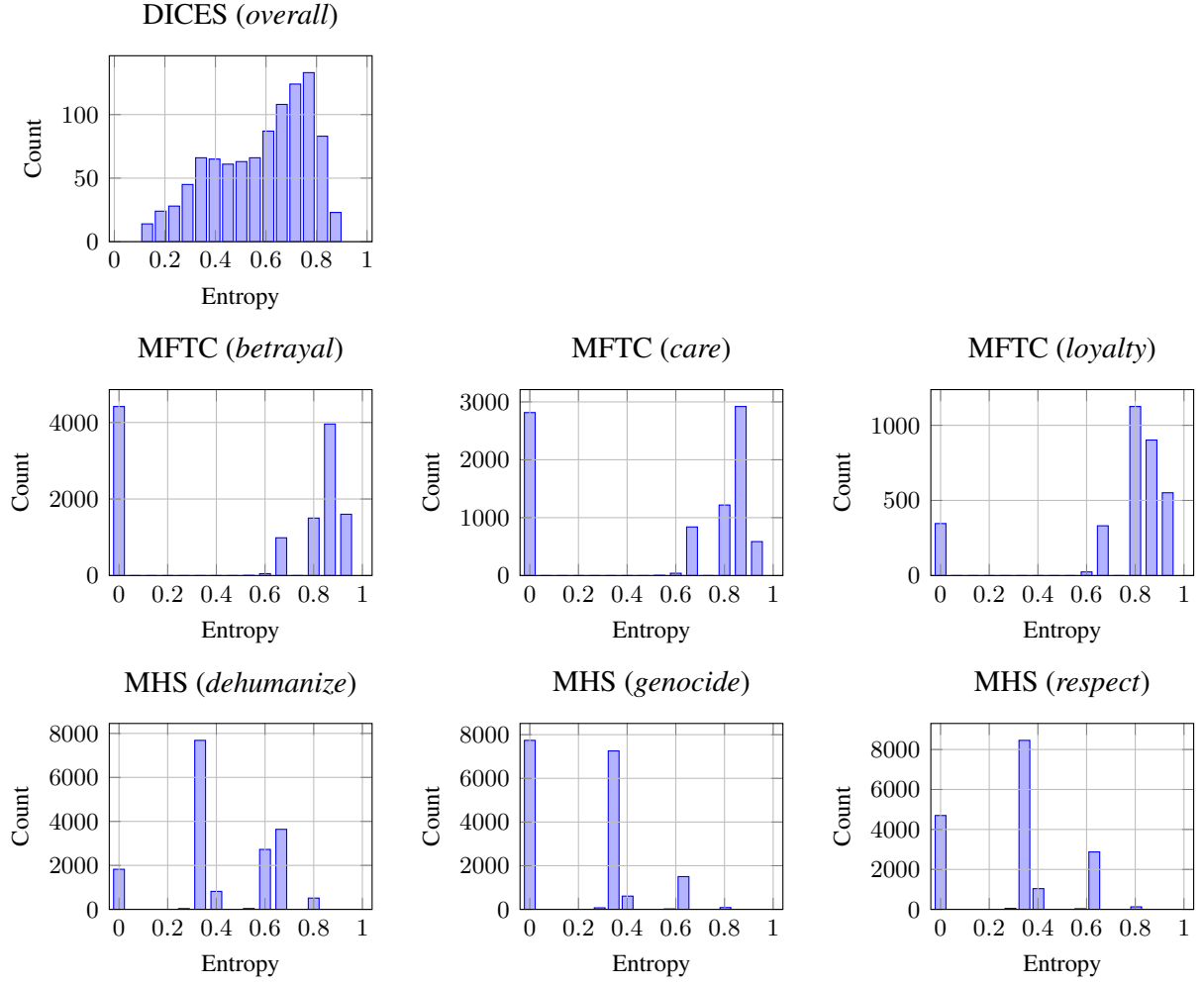
\begin{figure*}[ht]
    \centering
    \begin{tikzpicture}
    \begin{axis}[
        ymin=0, 
        name=entropy_dices_overall,
        title={DICES (\emph{overall})},
        xlabel={Entropy},
        ylabel={Count},
        bar width=4pt,
        xtick distance=0.2,
        height=4cm,
        width=5cm,
        xmin=0,
        xmax=1,
        grid,
        minor y tick num = 0,
        area style,
        enlarge x limits=0.02,
    ]
        \addplot+[ybar,mark=yes] plot coordinates { 
(0.13015, 14.00000)
(0.18342, 24.00000)
(0.23668, 28.00000)
(0.28995, 45.00000)
(0.34321, 66.00000)
(0.39648, 65.00000)
(0.44974, 61.00000)
(0.50301, 63.00000)
(0.55627, 66.00000)
(0.60954, 87.00000)
(0.66280, 108.00000)
(0.71607, 124.00000)
(0.76933, 133.00000)
(0.82260, 83.00000)
(0.87587, 23.00000)
        };
    \end{axis}


    \begin{axis}[
        ymin=0, 
        name=entropy_mftc_betrayal,
        at={($(entropy_dices_overall.south)+(0em, -5em)$)},
        anchor=north,
        title={MFTC (\emph{betrayal})},
        xlabel={Entropy},
        ylabel={Count},
        bar width=4pt,
        height=4cm,
        xtick distance=0.2,
        width=5cm,
        xmin=0,
        xmax=1,
        grid,
        minor y tick num = 0,
        area style,
        enlarge x limits=0.04,
    ]
        \addplot+[ybar] plot coordinates { 
(0.00000, 4417.00000)
(0.06667, 0.00000)
(0.13333, 0.00000)
(0.20000, 0.00000)
(0.26667, 0.00000)
(0.33333, 0.00000)
(0.40000, 0.00000)
(0.46667, 0.00000)
(0.53333, 1.00000)
(0.60000, 49.00000)
(0.66667, 986.00000)
(0.73333, 0.00000)
(0.80000, 1501.00000)
(0.86667, 3959.00000)
(0.93333, 1601.00000)
        };
    \end{axis}

    \begin{axis}[
        ymin=0, 
        name=entropy_mftc_care,
        at={($(entropy_mftc_betrayal.east)+(5em, 0em)$)},
        anchor=west,
        title={MFTC (\emph{care})},
        xlabel={Entropy},
        ylabel={Count},
        bar width=4pt,
        height=4cm,
        xtick distance=0.2,
        width=5cm,
        xmin=0,
        xmax=1,
        grid,
        minor y tick num = 0,
        area style,
        enlarge x limits=0.04,
    ]
        \addplot+[ybar] plot coordinates { 
(0.00000, 2815.00000)
(0.06667, 0.00000)
(0.13333, 0.00000)
(0.20000, 0.00000)
(0.26667, 0.00000)
(0.33333, 0.00000)
(0.40000, 0.00000)
(0.46667, 0.00000)
(0.53333, 2.00000)
(0.60000, 39.00000)
(0.66667, 839.00000)
(0.73333, 0.00000)
(0.80000, 1218.00000)
(0.86667, 2920.00000)
(0.93333, 586.00000)
        };
    \end{axis}

    \begin{axis}[
        ymin=0, 
        name=entropy_mftc_loyalty,
        at={($(entropy_mftc_care.east)+(5em, 0em)$)},
        anchor=west,
        title={MFTC (\emph{loyalty})},
        xlabel={Entropy},
        ylabel={Count},
        bar width=4pt,
        height=4cm,
        xtick distance=0.2,
        width=5cm,
        xmin=0,
        xmax=1,
        grid,
        minor y tick num = 0,
        area style,
        enlarge x limits=0.04,
    ]
        \addplot+[ybar] plot coordinates { 
(0.00000, 346.00000)
(0.06667, 0.00000)
(0.13333, 0.00000)
(0.20000, 0.00000)
(0.26667, 0.00000)
(0.33333, 0.00000)
(0.40000, 0.00000)
(0.46667, 0.00000)
(0.53333, 0.00000)
(0.60000, 24.00000)
(0.66667, 331.00000)
(0.73333, 0.00000)
(0.80000, 1125.00000)
(0.86667, 902.00000)
(0.93333, 551.00000)
        };
    \end{axis}

    \begin{axis}[
        ymin=0, 
        name=entropy_mhs_dehumanize,
        at={($(entropy_mftc_betrayal.south)+(0em, -5em)$)},
        anchor=north,
        title={MHS (\emph{dehumanize})},
        xlabel={Entropy},
        ylabel={Count},
        bar width=4pt,
        height=4cm,
        xtick distance=0.2,
        width=5cm,
        xmin=0,
        xmax=1,
        grid,
        minor y tick num = 0,
        area style,
        enlarge x limits=0.04,
    ]
        \addplot+[ybar] plot coordinates { 
(0.00000, 1823.00000)
(0.06667, 0.00000)
(0.13333, 0.00000)
(0.20000, 0.00000)
(0.26667, 37.00000)
(0.33333, 7684.00000)
(0.40000, 818.00000)
(0.46667, 0.00000)
(0.53333, 40.00000)
(0.60000, 2725.00000)
(0.66667, 3640.00000)
(0.73333, 0.00000)
(0.80000, 513.00000)
(0.86667, 0.00000)
(0.93333, 2.00000)
        };
    \end{axis}

    \begin{axis}[
        ymin=0, 
        name=entropy_mhs_genocide,
        at={($(entropy_mhs_dehumanize.east)+(5em, 0em)$)},
        anchor=west,
        title={MHS (\emph{genocide})},
        xlabel={Entropy},
        ylabel={Count},
        bar width=4pt,
        height=4cm,
        xtick distance=0.2,
        width=5cm,
        xmin=0,
        xmax=1,
        grid,
        minor y tick num = 0,
        area style,
        enlarge x limits=0.04,
    ]
        \addplot+[ybar] plot coordinates { 
(0.00000, 7739.00000)
(0.05742, 0.00000)
(0.11485, 0.00000)
(0.17227, 0.00000)
(0.22969, 0.00000)
(0.28712, 71.00000)
(0.34454, 7258.00000)
(0.40196, 607.00000)
(0.45939, 0.00000)
(0.51681, 0.00000)
(0.57424, 22.00000)
(0.63166, 1501.00000)
(0.68908, 0.00000)
(0.74651, 0.00000)
(0.80393, 84.00000)
        };
    \end{axis}

    \begin{axis}[
        ymin=0, 
        name=entropy_mhs_respect,
        at={($(entropy_mhs_genocide.east)+(5em, 0em)$)},
        anchor=west,
        title={MHS (\emph{respect})},
        xlabel={Entropy},
        ylabel={Count},
        bar width=4pt,
        height=4cm,
        xtick distance=0.2,
        width=5cm,
        xmin=0,
        xmax=1,
        grid,
        minor y tick num = 0,
        area style,
        enlarge x limits=0.04,
    ]
        \addplot+[ybar] plot coordinates { 
(0.00000, 4698.00000)
(0.05742, 0.00000)
(0.11485, 0.00000)
(0.17227, 0.00000)
(0.22969, 1.00000)
(0.28712, 50.00000)
(0.34454, 8458.00000)
(0.40196, 1038.00000)
(0.45939, 0.00000)
(0.51681, 0.00000)
(0.57424, 33.00000)
(0.63166, 2877.00000)
(0.68908, 0.00000)
(0.74651, 0.00000)
(0.80393, 127.00000)
        };
    \end{axis}
\end{tikzpicture}
    \caption{Histogram of entropy score over all annotations per sample for each dataset and task combination.}
    \label{app:fig:entropy-scores}
\end{figure*}

We report the average disagreement rates per dataset and task in Figure~\ref{app:fig:entropy-scores}, for each of the dataset and task combinations.

\renewcommand{\thefigure}{B\arabic{figure}}
\setcounter{figure}{0}
\renewcommand{\thetable}{B\arabic{table}}
\setcounter{table}{0}

\section{Detailed results overview}
\label{app:detailed-results}
\subsection{Annotator-Centric evaluation for other MFTC and MHS tasks}
We show the full annotator-centric metrics results for MFTC \emph{betrayal}, MFTC \emph{loyalty}, MHS \emph{genocide}, and MHS \emph{respect} in Table~\ref{tab:annotator-centric-other-tasks}. This follows the same format at Table~\ref{tab:annotator-centric-all}. The results in this table also form the basis for Figure~\ref{fig:task-comparison}.

\begin{table}[ht]
    \small
        \centering
        \begin{tabular}{@{}cl@{\hspace{.2cm}}c@{\hspace{.2cm}}c@{\hspace{.2cm}}c@{\hspace{.2cm}}c@{\hspace{.2cm}}c@{\hspace{.2cm}}c@{\hspace{.2cm}}c@{}}
             \toprule
             & & & & \multicolumn{2}{c}{\textbf{Average}} & \multicolumn{2}{c}{\textbf{Worst-off}}\\
             & \textbf{App.} & $F_1$ & $JS$ & $F_1^a $ &  $JS^a$ & $F_1^w$ & $JS^w$ & $\Delta\%$\\
             \midrule
    \parbox[t]{2mm}{\multirow{11}{*}{\rotatebox[origin=c]{90}{MFTC (\emph{betrayal})}}}
& $\mathcal{S}_R$$\mathcal{T}_R$    & 71.5 & .047 & 57.8 & \textbf{.147} & 42.0 & .199 & -1.6 \\
& $\mathcal{S}_R$$\mathcal{T}_L$    & 71.2 & .046 & 58.1 & .149 & 43.3 & .212 & -1.6 \\
& $\mathcal{S}_R$$\mathcal{T}_S$    & 71.2 & .051 & 59.3 & .161 & 43.0 & .239 & -5.0 \\
& $\mathcal{S}_R$$\mathcal{T}_D$    & 71.0 & .046 & 58.3 & .148 & 42.9 & .199 & -1.6 \\
& $\mathcal{S}_U$$\mathcal{T}_R$    & 72.6 & .042 & \textbf{59.4} & .150 & 41.9 & .203 & -2.5 \\
& $\mathcal{S}_U$$\mathcal{T}_L$    & 73.6 & .045 & 58.4 & .148 & 43.4 & .200 & -1.3 \\
& $\mathcal{S}_U$$\mathcal{T}_S$    & 74.0 & .045 & 58.8 & .149 & \textbf{43.5} & .204 & -1.0 \\
& $\mathcal{S}_U$$\mathcal{T}_D$    & 73.2 & .044 & 59.1 & .149 & 42.8 & \textbf{.194} & -2.6 \\
\cmidrule(ll){2-9}
& $\mathcal{S}_R$$\mathcal{O}$      & 72.1 & .046 & 58.9 & \textbf{.147} & 43.1 & .195 & \textbf{-48.6} \\
& $\mathcal{S}_U$$\mathcal{O}$      & 71.8 & .047 & 58.9 & .149 & 43.0 & .200 & -0.0 \\
& PL                                & \textbf{75.2} & \textbf{.037} & 48.1 & .199 & 36.0 & .290 & 0.0 \\
\midrule
\parbox[t]{2mm}{\multirow{11}{*}{\rotatebox[origin=c]{90}{MFTC (\emph{betrayal})}}}
& $\mathcal{S}_R$$\mathcal{T}_R$    & 66.9 & .034 & 56.4 & .177 & 22.2 & .372 & -0.4 \\
& $\mathcal{S}_R$$\mathcal{T}_L$    & 68.9 & .032 & 56.3 & .176 & 22.2 & .374 & -0.3 \\
& $\mathcal{S}_R$$\mathcal{T}_S$    & 67.1 & .031 & \textbf{57.3} & .176 & 22.2 & .370 & -0.3 \\
& $\mathcal{S}_R$$\mathcal{T}_D$    & 68.4 & .031 & 55.1 & \textbf{.175} & 22.2 & .373 & -0.3 \\
& $\mathcal{S}_U$$\mathcal{T}_R$    & 61.3 & .032 & 55.7 & .177 & 21.7 & .357 & -1.1 \\
& $\mathcal{S}_U$$\mathcal{T}_L$    & 66.5 & .032 & 54.1 & .177 & 22.2 & .355 & -0.8 \\
& $\mathcal{S}_U$$\mathcal{T}_S$    & 62.4 & .033 & 55.6 & .177 & 22.2 & .358 & -0.9 \\
& $\mathcal{S}_U$$\mathcal{T}_D$    & 64.4 & .031 & 55.8 & .177 & 22.2 & .358 & -1.3 \\
\cmidrule(ll){2-9}
& $\mathcal{S}_R$$\mathcal{O}$      & \textbf{71.5} & .030 & 56.0 & .176 & 22.2 & .361 & \textbf{-29.1} \\
& $\mathcal{S}_U$$\mathcal{O}$      & 66.5 & .033 & 55.9 & .177 & 22.2 & .366 & -0.1 \\
& PL                                & 62.5 & \textbf{.029} & 51.2 & .183 & \textbf{26.1} & \textbf{.309} & 0.0 \\
\midrule
\parbox[t]{2mm}{\multirow{11}{*}{\rotatebox[origin=c]{90}{MHS (\emph{genocide})}}}
& $\mathcal{S}_R$$\mathcal{T}_R$    & 26.5 & .050 & 70.0 & .227 & 0.0 & .560 & -6.3 \\
& $\mathcal{S}_R$$\mathcal{T}_L$    & 28.2 & .051 & 69.8 & .225 & 0.0 & .565 & -1.7 \\
& $\mathcal{S}_R$$\mathcal{T}_S$    & 28.1 & .051 & 70.0 & \textbf{.224} & 0.0 & .566 & -1.7 \\
& $\mathcal{S}_R$$\mathcal{T}_D$    & 28.3 & .050 & 70.2 & \textbf{.224} & 0.0 & .565 & -1.7 \\
& $\mathcal{S}_U$$\mathcal{T}_R$    & 32.8 & .077 & 71.1 & .229 & 0.0 & .549 & -12.6 \\
& $\mathcal{S}_U$$\mathcal{T}_L$    & 27.7 & .048 & 70.7 & .231 & 0.0 & .548 & -7.9 \\
& $\mathcal{S}_U$$\mathcal{T}_S$    & 26.7 & .048 & 70.9 & .231 & 0.0 & .548 & -7.9 \\
& $\mathcal{S}_U$$\mathcal{T}_D$    & 27.3 & .048 & \textbf{71.2} & .229 & 0.0 & .547 & -12.6 \\
\cmidrule(ll){2-9}
& $\mathcal{S}_R$$\mathcal{O}$      & 28.0 & .048 & 33.9 & .387 & 0.0 & \textbf{.496} & \textbf{-60.1} \\
& $\mathcal{S}_U$$\mathcal{O}$      & \textbf{33.3} & .080 & 33.1 & .390 & 0.0 & .497 & -24.7 \\
& PL                                & 21.6 & \textbf{.044} & 70.0 & .245 & 0.0 & .570 & -- \\
\midrule
\parbox[t]{2mm}{\multirow{11}{*}{\rotatebox[origin=c]{90}{MHS (\emph{respect})}}}
& $\mathcal{S}_R$$\mathcal{T}_R$    & 41.4 & .086 & 46.0 & .331 & 0.0 & .528 & -18.8 \\
& $\mathcal{S}_R$$\mathcal{T}_L$    & 40.8 & .087 & 45.6 & .331 & 0.0 & .530 & -18.8 \\
& $\mathcal{S}_R$$\mathcal{T}_S$    & 41.2 & .086 & 46.1 & .331 & 0.0 & .529 & -18.8 \\
& $\mathcal{S}_R$$\mathcal{T}_D$    & 40.6 & .086 & 46.0 & .331 & 0.0 & .528 & -18.8 \\
& $\mathcal{S}_U$$\mathcal{T}_R$    & 32.8 & \textbf{.077} & \textbf{46.6} & \textbf{.323} & 0.0 & .533 & -4.9 \\
& $\mathcal{S}_U$$\mathcal{T}_L$    & 41.0 & .085 & 46.3 & \textbf{.323} & 0.0 & .532 & -4.9 \\
& $\mathcal{S}_U$$\mathcal{T}_S$    & \textbf{41.8} & .084 & 45.9 & .324 & 0.0 & .531 & -4.9 \\
& $\mathcal{S}_U$$\mathcal{T}_D$    & 40.6 & .085 & 46.2 & .324 & 0.0 & .532 & -4.9 \\
\cmidrule(ll){2-9}
& $\mathcal{S}_R$$\mathcal{O}$      & 41.7 & .085 & 33.9 & .387 & 0.0 & \textbf{.496} & \textbf{-60.1} \\
& $\mathcal{S}_U$$\mathcal{O}$      & 33.3 & .080 & 33.1 & .390 & 0.0 & .497 & -24.7 \\
& PL                                & 41.0 & .080 & 25.9 & .405 & 0.0 & .587 & -- \\
\bottomrule
\end{tabular}
\caption{Test set results on the MFTC (\textit{betrayal}), MFTC (\textit{loyalty}), MHS (\textit{genocide}), and MHS (\textit{respect}) datasets. $\Delta\%$ denotes the reduction in the annotation budget with respect to passive learning.}
\label{tab:annotator-centric-other-tasks}
\end{table}

\subsection{Training process}
\label{app:detailed-results-training}
\begin{figure*}
    \centering
    \includestandalone[width=\textwidth]{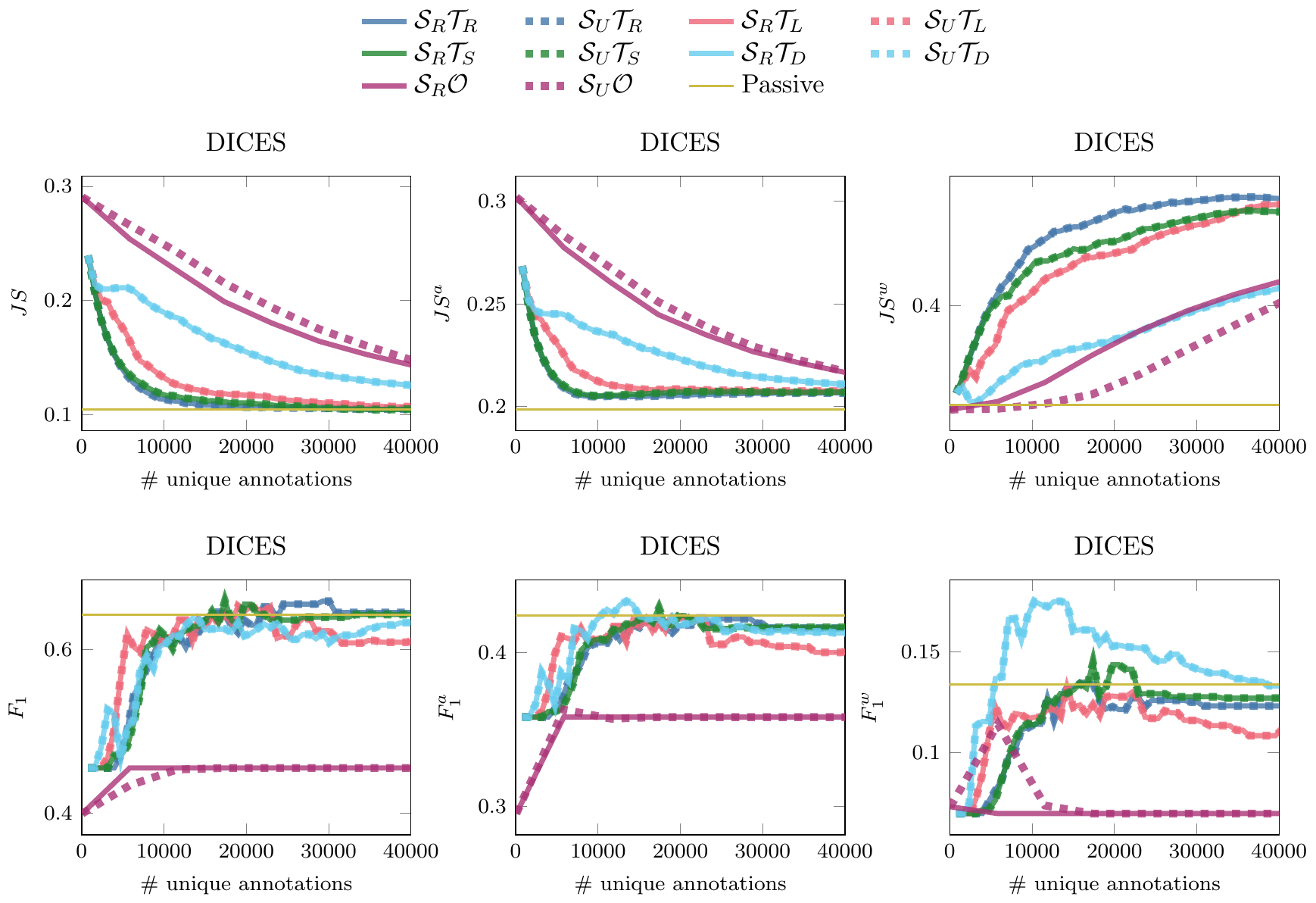}
    \caption{Validation set performance across all metrics for DICES during training.}
    \label{fig:progress-dices-overall}
\end{figure*}

\begin{figure*}
    \centering
    \includestandalone[width=\textwidth]{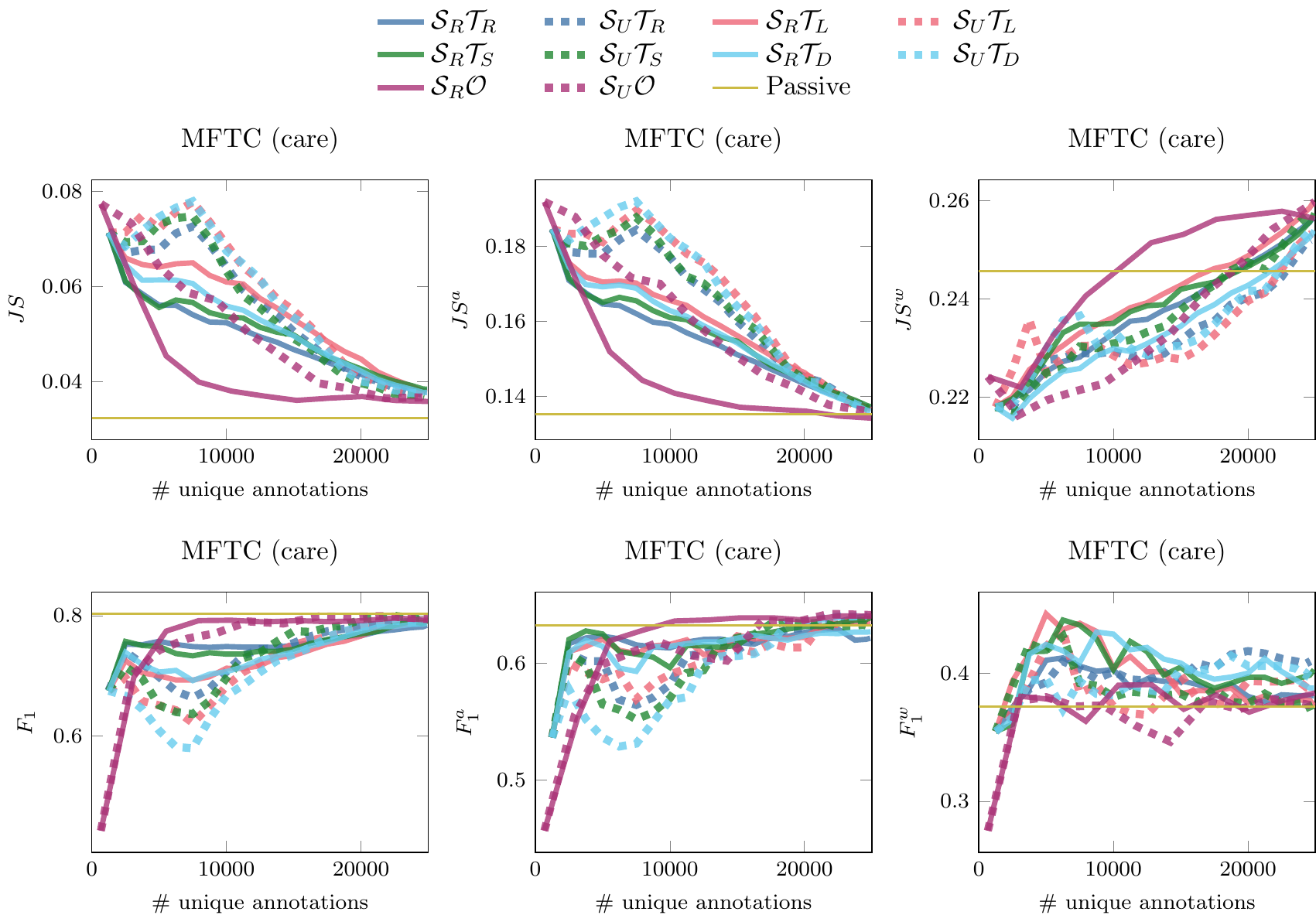}
    \caption{Validation set performance across all metrics for MFTC (care) during training}
    \label{fig:progress-mftc-care}
\end{figure*}

\begin{figure*}
    \centering
    \includestandalone[width=\textwidth]{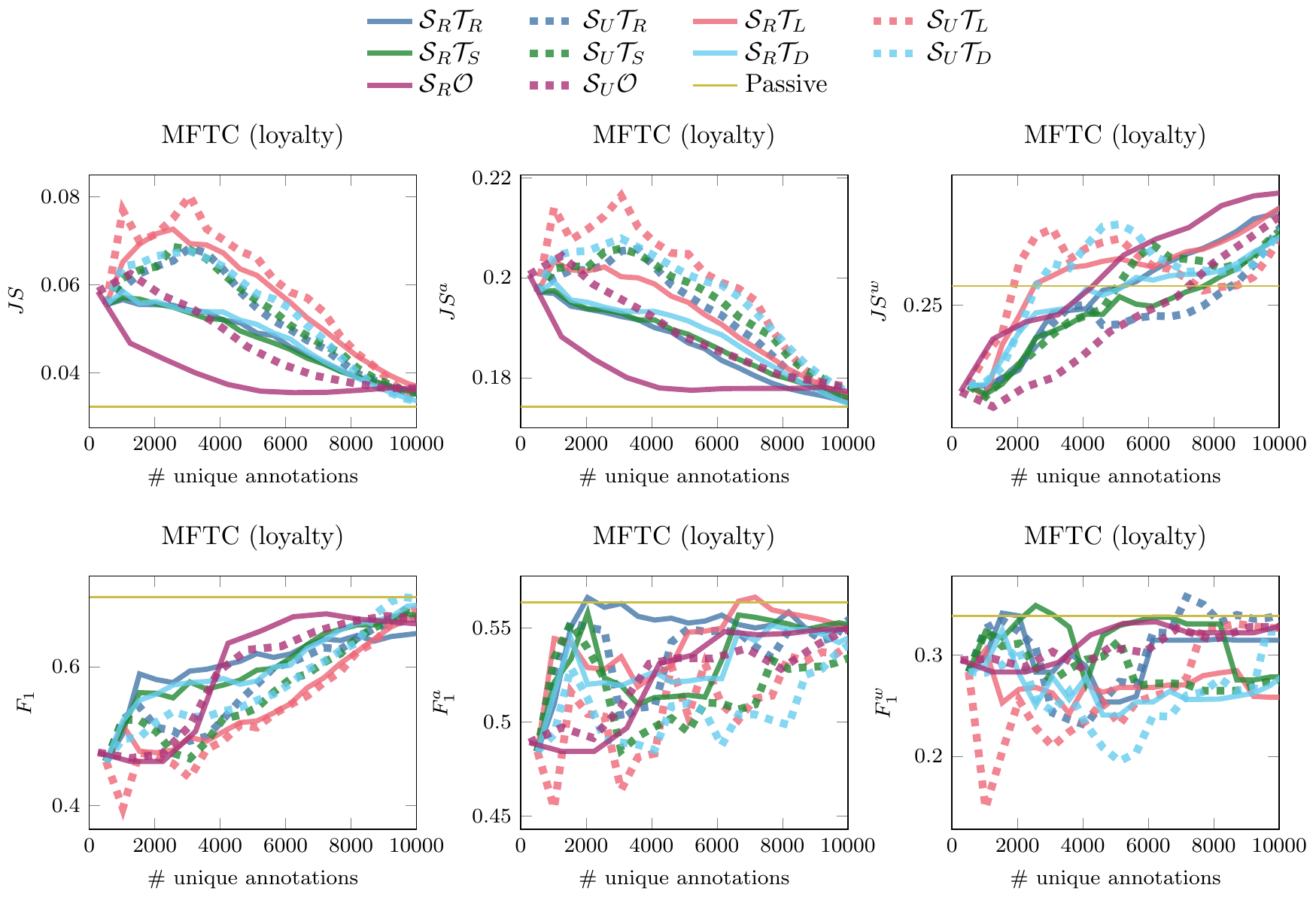}
    \caption{Validation set performance across all metrics for MFTC (loyalty) during training}
    \label{fig:progress-mftc-loyalty}
\end{figure*}

\begin{figure*}
    \centering
    \includestandalone[width=\textwidth]{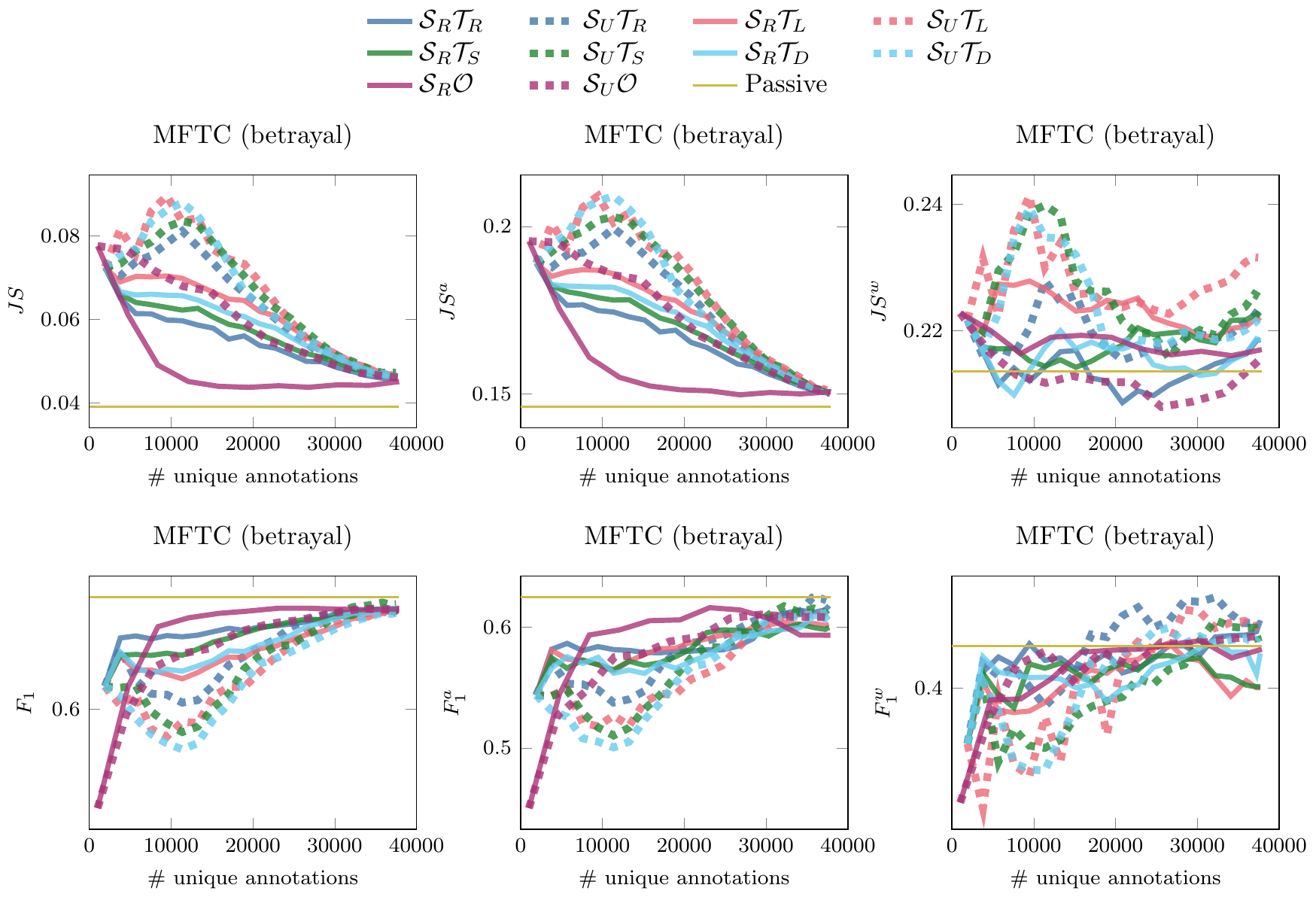}
    \caption{Validation set performance across all metrics for MFTC (betrayal) during training}
    \label{fig:progress-mftc-betrayal}
\end{figure*}

\begin{figure*}
    \centering
    \includestandalone[width=\textwidth]{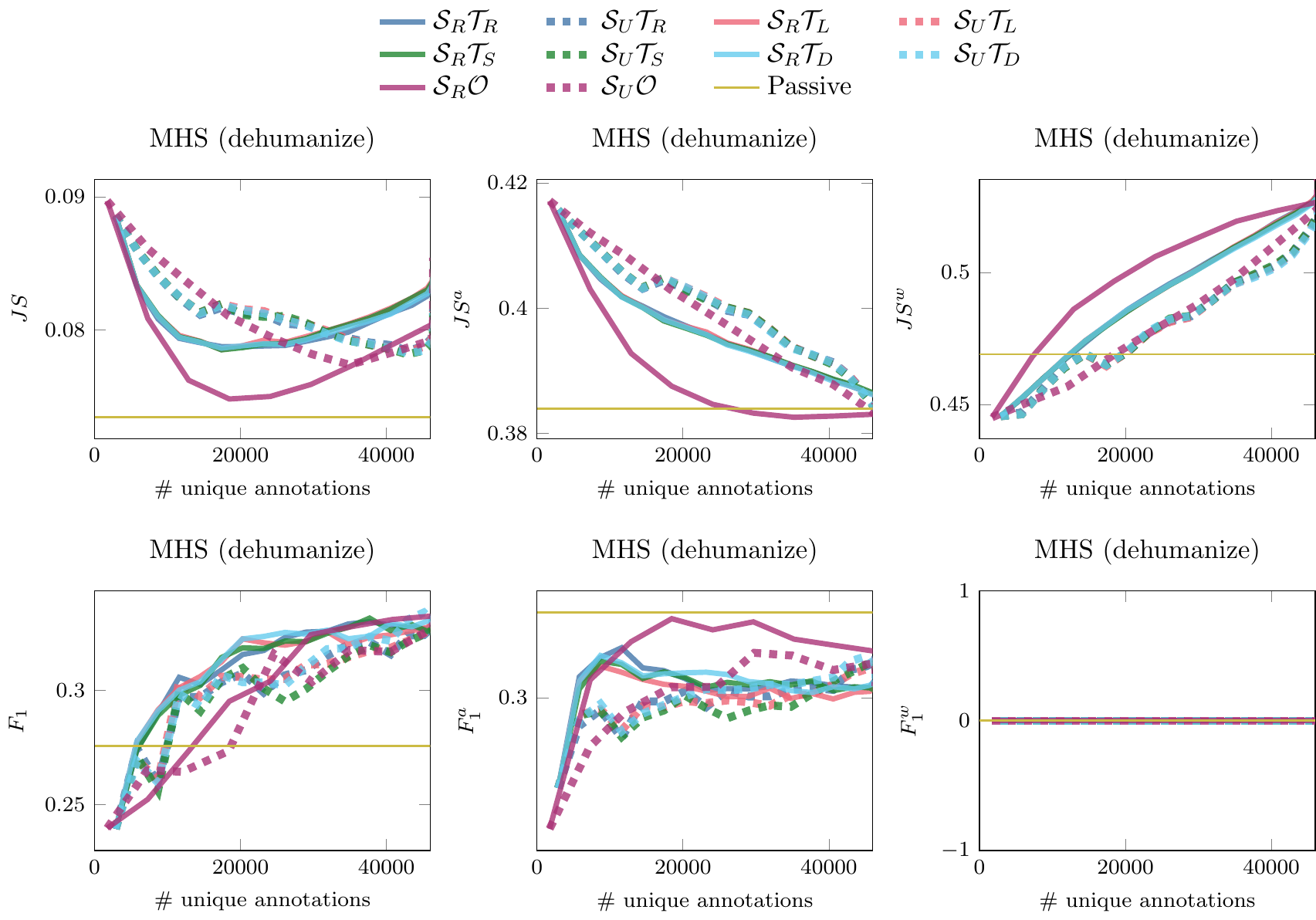}
    \caption{Validation set performance across all metrics for MHS (dehumanize) during training}
    \label{fig:progress-mhs-dehumanize}
\end{figure*}

\begin{figure*}
    \centering
    \includestandalone[width=\textwidth]{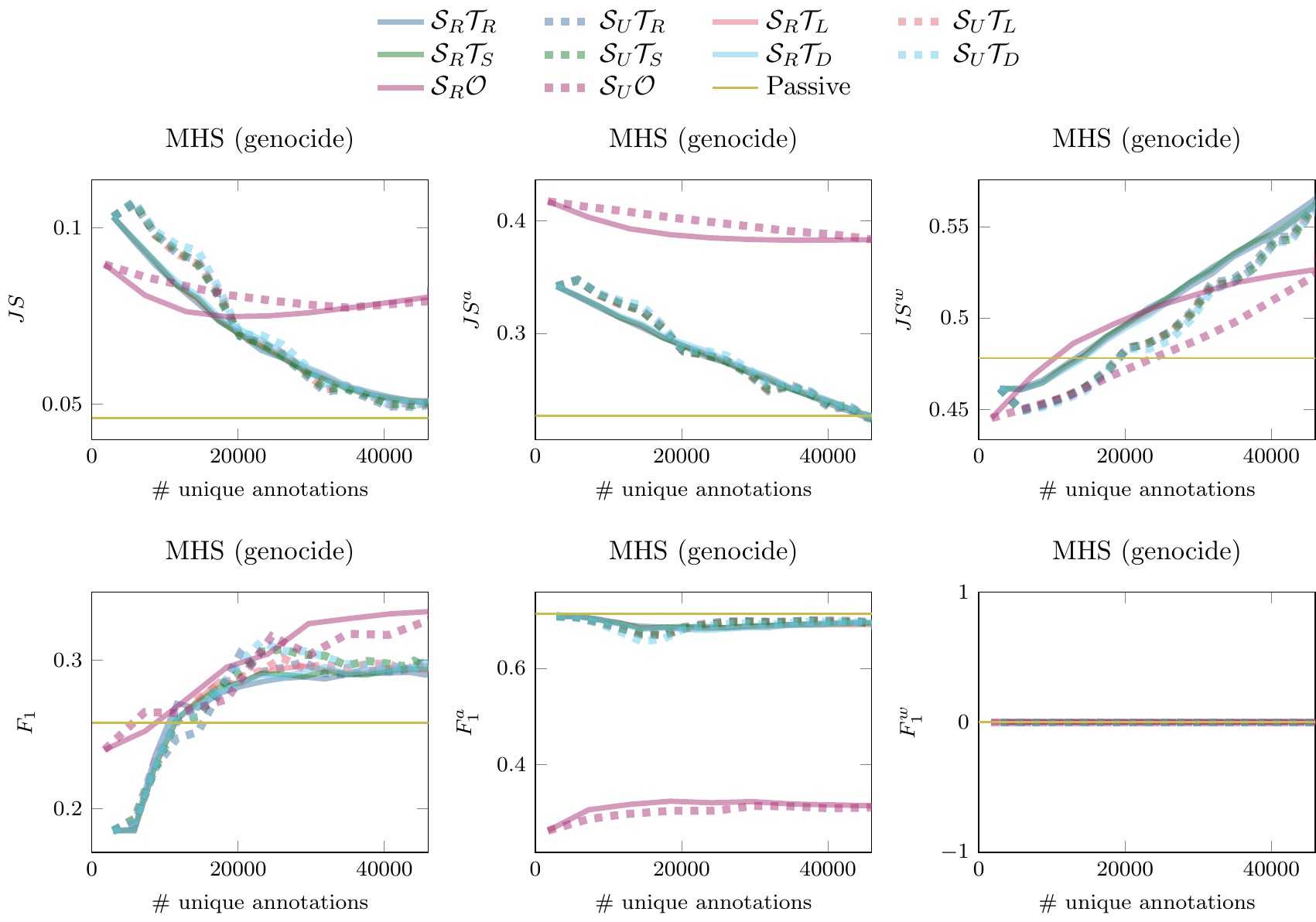}
    \caption{Validation set performance across all metrics for MHS (genocide) during training}
    \label{fig:progress-mhs-genocide}
\end{figure*}

\begin{figure*}
    \centering
    \includestandalone[width=\textwidth]{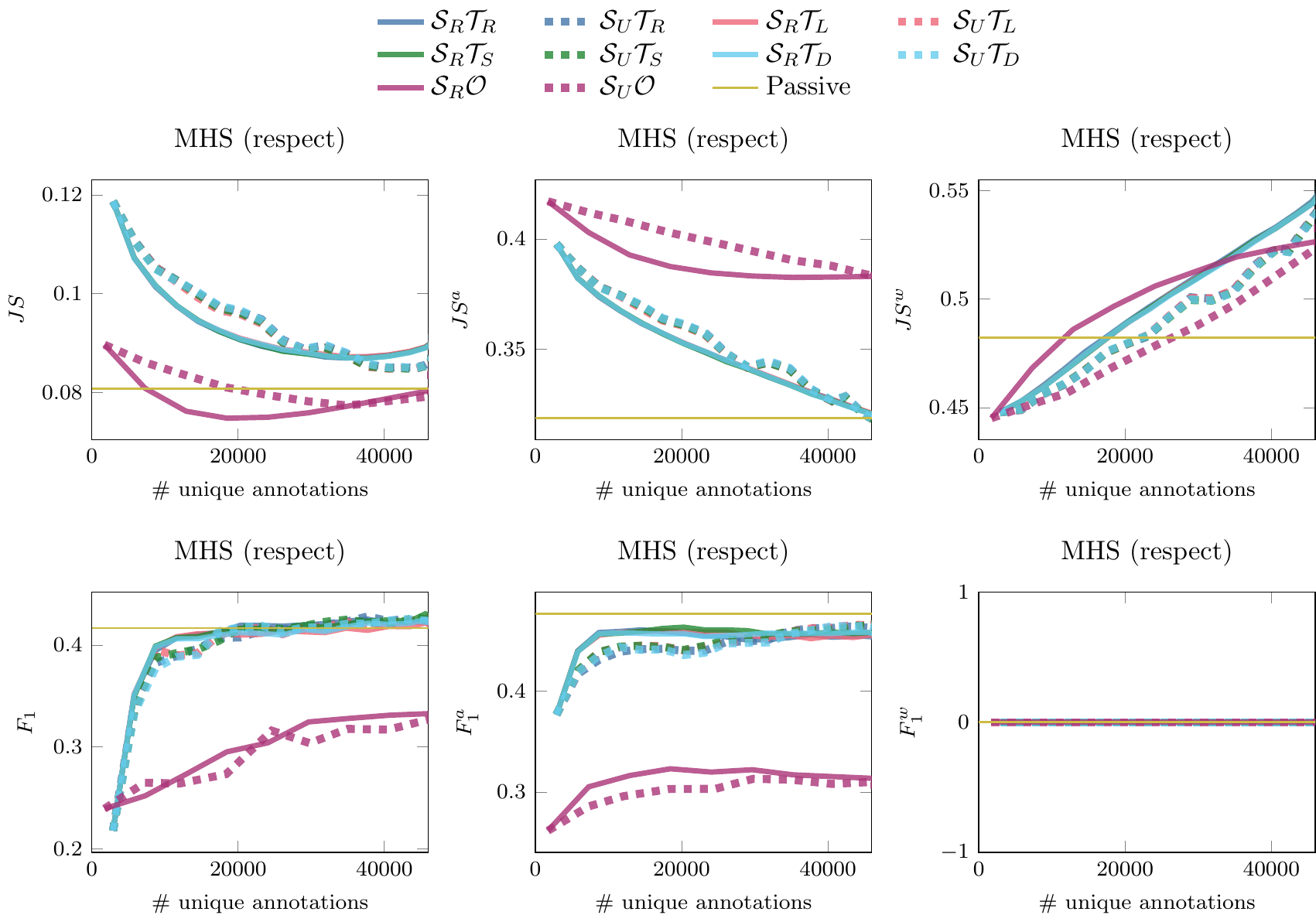}
    \caption{Validation set performance across all metrics for MHS (respect) during training}
    \label{fig:progress-mhs-respect}
\end{figure*}

In our main paper, we report a condensed version of all metrics during the training phase of the active learning approaches. Below, we provide a complete overview of all approaches over all metrics. The results can be seen in Figures~\ref{fig:progress-dices-overall} through~\ref{fig:progress-mhs-respect}.

\renewcommand{\tfigwidth}{8cm}
\renewcommand{\tfigheight}{5cm}
\renewcommand{\customlinewidth}{2pt}
\renewcommand{\customlinewidthdotted}{3pt}
\renewcommand{\customlegendxoffset}{11cm}
\renewcommand{\customlegendyoffset}{4.66cm}
\renewcommand{\lineopacity}{0.8}
\renewcommand{\custompassivelinewidth}{0.8pt}

\end{document}